%% file: iclr2027_conference.tex
\documentclass{article} 
\usepackage{iclr2027_conference,times}

\input{math_commands.tex}

\usepackage{hyperref}
\usepackage{float}
\usepackage{url}
\usepackage{graphicx}

\title{Layer-Wise Input Recoverability in Deep Neural Networks}

\author{Max Spiridonov \\
Department of Computer Science\\
University of California, Irvine\\
\texttt{mspirido@uci.edu} \\
\And
Alireza Rahmansetayesh \\
Department of Computer Science\\
University of California, Irvine\\
\texttt{arahman@uci.edu} \\
\And
Pierre Baldi \\
Department of Computer Science\\
University of California, Irvine\\
\texttt{pfbaldi@ics.uci.edu} \\
}

\iclrfinalcopy 
\begin{document}

\maketitle

\begin{abstract}
The inner workings of deep neural networks are still not well understood. The information bottleneck framework, which characterizes deep networks by the mutual information between their hidden representations and the network's input, has been used to identify compression of input information with respect to an input image and output labels. We instead study how much of a layer's own input is recoverable from the layer's output in convolutional neural networks (CNNs). A convolutional layer applies its filters to one input patch at a time, so we evaluate recoverability on sampled input patches with a reconstruction-based probe and not on the entire feature map. The probe's size thus depends on channels and kernel size and not on feature map size, which lets our method scale to deep CNNs with wide feature maps. For each layer, we compare pretrained, random, and reconstruction-optimized encoders, using linear and linear-plus-residual-MLP decoders. We measure recoverability as $R^2$ between input patches and their reconstructions on a held-out set. Across nine ImageNet CNNs, recoverability generally decreases with depth but is more consistently predicted by the compression of the layer. Learning-attributed recoverability, the difference between pretrained and random encoders, is not uniform across architectures. We introduce single-property random controls, which transfer one pretrained property onto a random encoder, and find that each control changes recoverability with a consistent sign across models: copying pretrained singular vectors tends to raise recoverability, while copying the pretrained singular-value spectrum or activation sparsity lowers it.
\end{abstract}

\section{Introduction}

Deep networks transform their input through a sequence of layers, and a long-standing question is how much information about the input each layer preserves. One motivation comes from the brain, which compresses visual information as it flows from the retina to higher cortical areas; humans recognize only about 40 bits of visual information per second \citep{1056814}, from a stream of about one megabyte per second \citep{zhaoping2025}. The information bottleneck (IB) framework \citep{tishby2015deeplearninginformationbottleneck} asks the same question of deep neural networks by characterizing each layer through its mutual information with the network's input and the output.

Measuring this in modern deep networks has not been simple. Mutual information is difficult to estimate for deep-layer feature maps with high dimensions, and results depend heavily on the estimator \citep{9468892}. Early analyses were limited to small fully-connected networks \citep{shwartzziv2017openingblackboxdeep}, whereas \citet{Wickstrm2023} uses a kernel-based estimator to scale the framework to deep CNNs, measuring mutual information between a layer's representations and the network's input. We instead are concerned with the information between a layer's input and its output. Other work in representation visualization asks what a representation learns by mapping it back to the pixel domain, with deconvolutional, optimization-based, or up-convolutional inversions of deep features \citep{zeiler2013visualizingunderstandingconvolutionalnetworks,mahendran2014understandingdeepimagerepresentations,dosovitskiy2016invertingvisualrepresentationsconvolutional}. Probes trained to predict labels from a layer's representations \citep{alain2018understandingintermediatelayersusing} avoid reconstruction, but their results reflect the probe's capacity as well as the quality of the representations \citep{belinkov2021probingclassifierspromisesshortcomings}.

We ask a local question: at each layer, how much of the layer's own input can be recovered from its output? For each convolutional layer we sample the input patches it reads, $X\in\mathbb{R}^{d}$ with $d=C_{in}\cdot K^{2}$, encode them with the layer's weights, normalization, and activation, and fit a decoder to reconstruct $X$. We define recoverability as the held-out $R^{2}=1-\mathrm{MSE}(X,\hat X)/\mathrm{Var}(X)$, using a linear decoder and a linear decoder with a residual MLP of width $m\cdot C_{out}$, $m\in\{1,4,16\}$. Because our decoders are restricted, recoverability is closer to a measure of V-information \citep{xu2020theoryusableinformationcomputational} than to Shannon information; we assess the relationship between recoverability and Shannon information in Appendix~\ref{app:simple}.

Using input patches and their representations has two advantages: a layer's filters applied to a patch give exactly the convolution output at that location, so the operation is equivalent. Also, neighboring patches are highly redundant \citep{8594568}, so we do not need to use all patches from an input feature map. As a result, our method is computationally efficient, as our probes scale with channels and kernel size, not feature map size. 

A layer's recoverability reflects both what its architecture forces (for example, a non-injective activation or a narrow bottleneck) and what training adds. To separate the two, we compare a layer's pretrained encoder with a random encoder of the same shape, a common reference in probing, and define learning-attributed recoverability as $R^{2}_{pre}-R^{2}_{rand}$. A random encoder differs from the pretrained encoder in many ways at once, so this gap alone says little about what training changed. We therefore introduce single-property random controls, which transfer one property of the pretrained weights, like their singular vectors, singular-value spectrum, activation sparsity, or BatchNorm parameters, onto a random encoder. We find that each single-property random control changes recoverability in a consistent direction across models. For example, recoverability increases compared to random when copying the singular vectors of pretrained weights in 11 of 13 models.

Our contributions are:
\begin{enumerate}
 \item \textbf{Probing patch-level representations.} We probe each convolutional layer on its input patches instead of the full feature map, which allows our method to easily scale to deep CNNs with large feature maps. We find that sampling only a few patches per feature map gives similar results to using many.
 \item \textbf{Learning-attributed recoverability.} We measure the recoverability of a pretrained layer's representations compared to a random layer's representations of the same patches to separate the effect of learning from architecture. We see that the difference is not uniform across models.
 \item \textbf{Single-property random controls.} We transfer one property of the pretrained weights onto randomly initialized weights, like the pretrained singular vectors, singular-value spectrum, activation sparsity, or BatchNorm parameters. We see that each property changes recoverability in a consistent direction compared to the random baseline across models.
\end{enumerate}

\section{Related work}
\subsection{Information bottleneck theory of Deep Learning}
Deep networks were first analyzed through the lens of the information bottleneck (IB) by \citet{tishby2015deeplearninginformationbottleneck}. They suggested that any deep network can be quantified by the mutual information between its layers' hidden representations and the input and output variables. Furthermore, they stated that deep networks under supervised training learn to maximize mutual information between their hidden representation $\hat{X}$ and the output variable $Y$, with $\hat{X}$ being a maximally compressed representation of the input variable $X$.

\citet{shwartzziv2017openingblackboxdeep} took this framework and applied it to small (up to 7 layer) fully-connected neural networks on a simple binary classification task. They found two distinct training phases: a fitting phase where the information $I(\hat{X};Y)$ increases, and a longer compression phase where the information $I(X;\hat{X})$ decreases.

Several studies further applied this framework to different neural networks and with different mutual information estimators. \citet{michael2018on} performed experiments with different mutual information binning methods and activation functions on the MNIST dataset, and they found no compression phase during training when using the ReLU activation function. On the other hand, \citet{Gabri__2019} found compression can happen during learning even with ReLU, given that the task contains task-relevant and irrelevant information. \citet{yu2020understandingconvolutionalneuralnetworks} used a multivariate mutual information estimator in order to apply the framework to convolutional neural networks (CNNs), but they mention that their MI estimator did not scale to larger network widths like VGG or ResNet architectures.

Most relevant to our work, \citet{Cheng2018} applies the IB framework to 6-layer CNNs and the CIFAR-10 dataset, but for simplicity they only track information of the last layer representations. \citet{e25060899} performs the first IB analysis of VGG16 on the CIFAR-10 dataset, showing a fitting and compression phase during training. They additionally find that for VGG16, the data processing inequality ($I(X;Y) \geq I(X;Z) \text{ for a Markov chain } X \rightarrow Y \rightarrow Z$) holds for all but one layer. A meta-analysis of these studies shows that the choice of MI estimator has a profound effect on the results \citep{9468892}.

\subsection{Probing}
Probing is a widely used technique for understanding hidden layer representations in neural networks, allowing properties of deep networks to be studied without changing their behavior. \citet{alain2018understandingintermediatelayersusing} train a linear classifier on each layer of deep CNNs, finding that linear separability of the class label increases with depth. A second line of work probes by reconstruction: deconvolutional networks \citep{zeiler2013visualizingunderstandingconvolutionalnetworks}, optimization-based inversion \citep{mahendran2014understandingdeepimagerepresentations}, and up-convolutional networks \citep{dosovitskiy2016invertingvisualrepresentationsconvolutional} map the features of a deep layer back to the pixel domain to show what the layer retains, with shallow layers preserving photographic detail and deeper layers preserving progressively more invariant content. Because they reconstruct the whole image, the result mixes the contribution of every intervening layer with the inverse's own prior over natural images, and it is typically examined qualitatively for a few networks.

Probing raises an identifiability problem: probe accuracy reflects the probe's capacity as well as the information in the representation \citep{belinkov2021probingclassifierspromisesshortcomings}, and the field disagrees on the remedy. Some argue for minimal probes so that success is attributable to the representation, as with linear classifiers \citep{alain2018understandingintermediatelayersusing}, while others argue for the most expressive probe available, since a probe that is too weak underestimates the information \citep{pimentel2020informationtheoreticprobinglinguisticstructure}.

Our work is a reconstruction probe with a different target and an explicit treatment of probe capacity. First, the target is a layer's own input, read patch by patch, and not a label or the whole image: no labels are needed, the decoder inverts only the operation at that layer and not the layers before it, and the recoverability can therefore be attributed to that layer. Second, we address the capacity question on both sides. Our primary decoder is a linear one solved in closed form, with no optimizer to confound it, and a ladder of wider linear-plus-residual-MLP decoders that start at the linear solution shows how much more a larger decoder can recover, with a shuffled-target control that measures how much a decoder can fit without structure in the data (Appendix~\ref{app:capacity}). We do not claim that the result is an estimate of Shannon information but treat it as a decoder-relative quantity, in the sense of V-information \citep{xu2020theoryusableinformationcomputational}. Third, we use random encoders as the reference, as probing studies do to check that a probe reads structure from the representation and not from the probe itself, and we extend the idea with single-property random controls that keep one property of the pretrained weights and randomize the rest, and with an encoder trained to reconstruct the patches as a reference.

\section{Method}

\begin{figure}[t]
  \centering
  \includegraphics[width=0.85\linewidth]{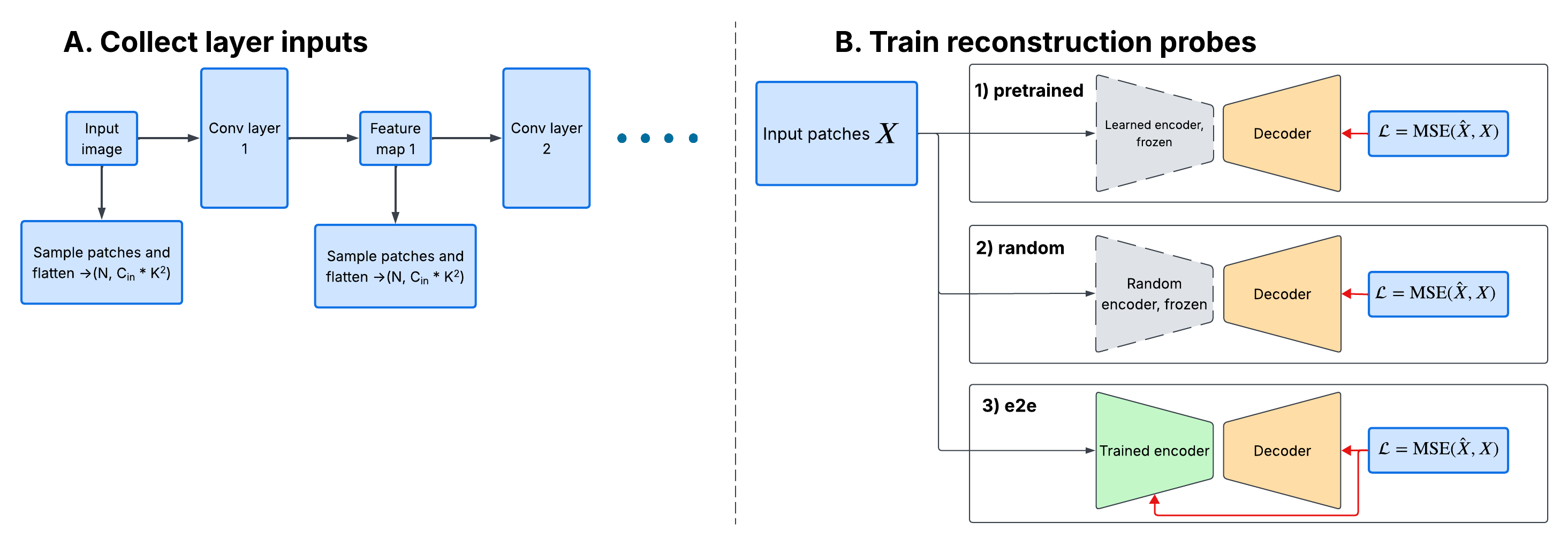}
   \caption{Method overview. \textbf{A.} Patches are sampled from each probed layer's input
 feature map (Section~\ref{sec:patches}). \textbf{B.} Each layer's patches are scored under
 three encodings of the same architecture: \textbf{1)} the network's pretrained
 encoder,
 \textbf{2)} a random encoder, and \textbf{3)} an encoder trained jointly with the decoder. All decoders are trained to minimize reconstruction mean squared error (MSE).}
  \label{fig:overview}
 \end{figure}
 
 \subsection{Sampling input patches}
 \label{sec:patches}

 We first perform a forward pass over our CNN. At each convolutional layer, we sample spatial positions from its input feature
 map (or image for the first layer). At each sampled position we extract the
 $K \times K$, $C_{\text{in}}$-channel patch the layer's filters read, flatten it to a vector
 $x \in \mathbb{R}^{d}$, where $d = C_{\text{in}} \cdot K^2$, and collect these patches into a
 tensor $X \in \mathbb{R}^{N \times d}$ per layer, where $N$ is the number of patches sampled. We used $N = 40{,}000 \text{ train and } 8{,}000$ validation images with 1 patch per image for our experiments, with ablations on number of images and patches per image in Appendix~\ref{app:patches}, where recoverability changes by at most 0.0104 and learning-attributed recoverability by at most 0.002.

 \subsection{Reconstruction probe}
For each layer's input patches, we first find their representations $z$. Our encoder is a fully-connected layer, followed by the layer's batch normalization and activation function. Thus, the representation $z$ of an input patch is exactly equal to the convolutional layer's output at that location. Now, our goal is to probe how recoverable a layer's input patches are given their representations.

 We fit a decoder $g^{(m)}$ reconstructing layer $m$'s input X from the patch representations. Our simplest decoder is a single fully-connected layer with no activation function---$\hat{X} = W_{lin}z + b_{lin}$. This allows us to perform convex optimization using linear least-squares, given patch representations $z$ and reconstruction targets $X$. The benefit of using a linear decoder is that the optimization has a closed-form solution and we do not need to use SGD. Our more expressive decoder adds a small nonlinear residual to the linear map. For a patch representation $z\in\mathbb{R}^{C_{out}}$ and an expansion
 factor $m\in\{1,4,16\}$,
 \begin{equation}
 \hat{X}=g^{(m)}(z)=\underbrace{W_{\mathrm{lin}}\,z+b_{\mathrm{lin}}}_{\text{linear part}}
 +\underbrace{W_{2}\,\sigma\!\left(W_{1}z+b_{1}\right)+b_{2}}_{\text{residual MLP}},
 \end{equation}
 where $W_{1}\in\mathbb{R}^{mC_{out}\times C_{out}}$, $W_{2}\in\mathbb{R}^{d\times mC_{out}}$, and $\sigma$ is the GELU activation function. We initialize
 $(W_{\mathrm{lin}},b_{\mathrm{lin}})$ at the closed-form solution and set $W_{2}=0$, $b_{2}=0$, so $g^{(m)}$ equals the linear decoder at
 initialization. We then train all parameters to minimize reconstruction mean-squared-error (MSE) with AdamW and early stopping to prevent overfitting. Appendix~\ref{app:capacity} evaluates how much recoverability depends on decoder capacity: for layers with a real bottleneck, wider decoders keep improving recoverability, so we read it as a lower bound, but the sign of the pretrained-minus-random difference is unchanged between the linear decoder and the widest decoder in 18 of 20 layer-model pairs.
 
 \subsection{Definition of Recoverability}
 \label{sec:definition}
 
 Recoverability is simply the coefficient of determination, $R^2 = 1 - \frac{\mathrm{MSE}\bigl(X,\; \hat{X}\bigr)}{\mathrm{Var}(X)}$
 for input patches $X$ and the reconstructed patches, $\hat{X}$, measured on validation set patches after probe training. $\mathrm{Var}(X)$ is the variance of the input patches $X$. Recoverability is thus the fraction of variance of a layer's input patches that can be reconstructed from their representations. Because our decoders are restricted, recoverability is a decoder-relative, V-information-style measure and not Shannon information. In simple examples where the information is directly computable, it tracks information when the decoder is optimal (rank correlation 0.97) and less closely for restricted decoders, and it weights the input's high-variance directions more than information does (Appendix~\ref{app:simple}).

 At each layer, we find the recoverability of input patches using the layer's pretrained weights. However, some recoverability is forced by architecture. For example, the commonly used ReLU activation function is non-injective, resulting in variance that is unable to be recovered. Layers with a higher compression ratio, measured by $d/C_{out}$, have a tighter bottleneck which also results in less recoverability. 
 To assess how pretrained weights affect recoverability, we compare the pretrained encoder against controls. Each
 control changes only the encoder weights of the probed layer; the input patches are identical across conditions, and
 the decoder is always optimized to minimize reconstruction MSE. We use three baseline encoders:

 \begin{enumerate}
  \item \textbf{\texttt{pretrained (pre)}}: the encoder is initialized and fixed with pretrained weights.
  \item \textbf{\texttt{random (rand)}}: the encoder is initialized with the framework's default random initialization
  and fixed, with the layer's BatchNorm (if any) re-estimated on training patches. Random encoders are a common
  reference for probes. We offer further analysis of our random controls in Appendix~\ref{app:controls}.
  \item \textbf{\texttt{end-to-end (e2e)}}: the encoder is initialized randomly and trained jointly with the decoder
  to minimize reconstruction MSE. This gives an optimization-limited reference of recoverability.
 \end{enumerate}

 We define learning-attributed recoverability, or the gap, as $R^2_{pre}-R^2_{rand}$. When it is positive, pretrained weights
 recover more of the input than random weights.

 To ask which properties of the pretrained weights produce this difference, we use single-property controls. Each
 starts from the default random encoder and replaces exactly one property with its pretrained value:

 \begin{enumerate}
  \item \textbf{\texttt{spectrum-matched (specm)}}: random singular vectors with the pretrained weights' singular
  values, $W=U_{\mathrm{rand}}\Sigma_{\mathrm{pre}}V_{\mathrm{rand}}^{\top}$ (excludes depthwise layers).
  \item \textbf{\texttt{singular-vector-matched (direcm)}}: the pretrained weights' singular vectors with the singular
   values of a default random draw, assigned in decreasing order,
  $W=U_{\mathrm{pre}}\Sigma_{\mathrm{rand}}V_{\mathrm{pre}}^{\top}$ (dense layers only).
  \item \textbf{\texttt{activation-sparsity-matched (sparsem)}}: random weights with each channel's bias shifted so that the
  channel's activation sparsity (firing rate) matches the pretrained
  channel's (layers with a nonlinearity only).
  \item \textbf{\texttt{BatchNorm-matched (bnm)}}: random weights with the pretrained BatchNorm scale and shift
  $(\gamma,\beta)$ (BatchNorm layers only).
 \end{enumerate}

 The effect of a property is $R^2_{control}-R^2_{rand}$, and the net effect of training is $R^2_{pre}-R^2_{rand}$. We say a control \emph{explains} the gap of a model (or layer type) if
  \begin{equation}
  \label{eq:explains}
  \frac{\sum_{l}\bigl(R^2_{pre}(l)-R^2_{control}(l)\bigr)^2}{\sum_{l}\bigl(R^2_{pre}(l)-R^2_{rand}(l)\bigr)^2}\;\le\;0.5,
  \end{equation}
  where $l$ indexes the layers of the model (or layer type) and the root-mean-square of $R^2_{pre}-R^2_{rand}$ over these layers is at least $0.02$ (a rule we fixed before inspecting the results). Explaining the gap does not mean that the property is a direct cause of recoverability, but rather shows how recoverability is associated with the property. Appendix~\ref{app:controls} gives the scores of all controls.

  In our experiments, we probe nine CNNs pretrained on ImageNet: VGG16, ResNet18, ResNet50, GoogLeNet, MobileNetV2, SqueezeNet 1.1, AlexNet, DenseNet121, and EfficientNet-B0. We use torchvision pretrained weights using ImageNet validation images at $224\times224$ resolution, and for VGG16 and ResNet18 that we trained on CIFAR-10 and CIFAR-100, we use the native $32\times32$ images. We probe every convolutional layer, including $1\times1$ and depthwise convolutions, except for skip-connection projections. We use 40,000 training and 8,000 validation images with one patch per image at every probed layer and the linear decoder. Results are means over probe seeds, which change the sampled patches and the random draws: 5 seeds for the ImageNet models and for VGG16 and ResNet18 on CIFAR-10, and 3 seeds for CIFAR-100 and for all controls. Appendix~\ref{app:patches} shows that other patch budgets give the same results, and Appendix~\ref{app:capacity} evaluates wider decoders.

\section{Results}

\subsection{Recoverability decreases over depth in most models}
\begin{figure}[!ht]
  \centering
  \includegraphics[width=0.85\linewidth]{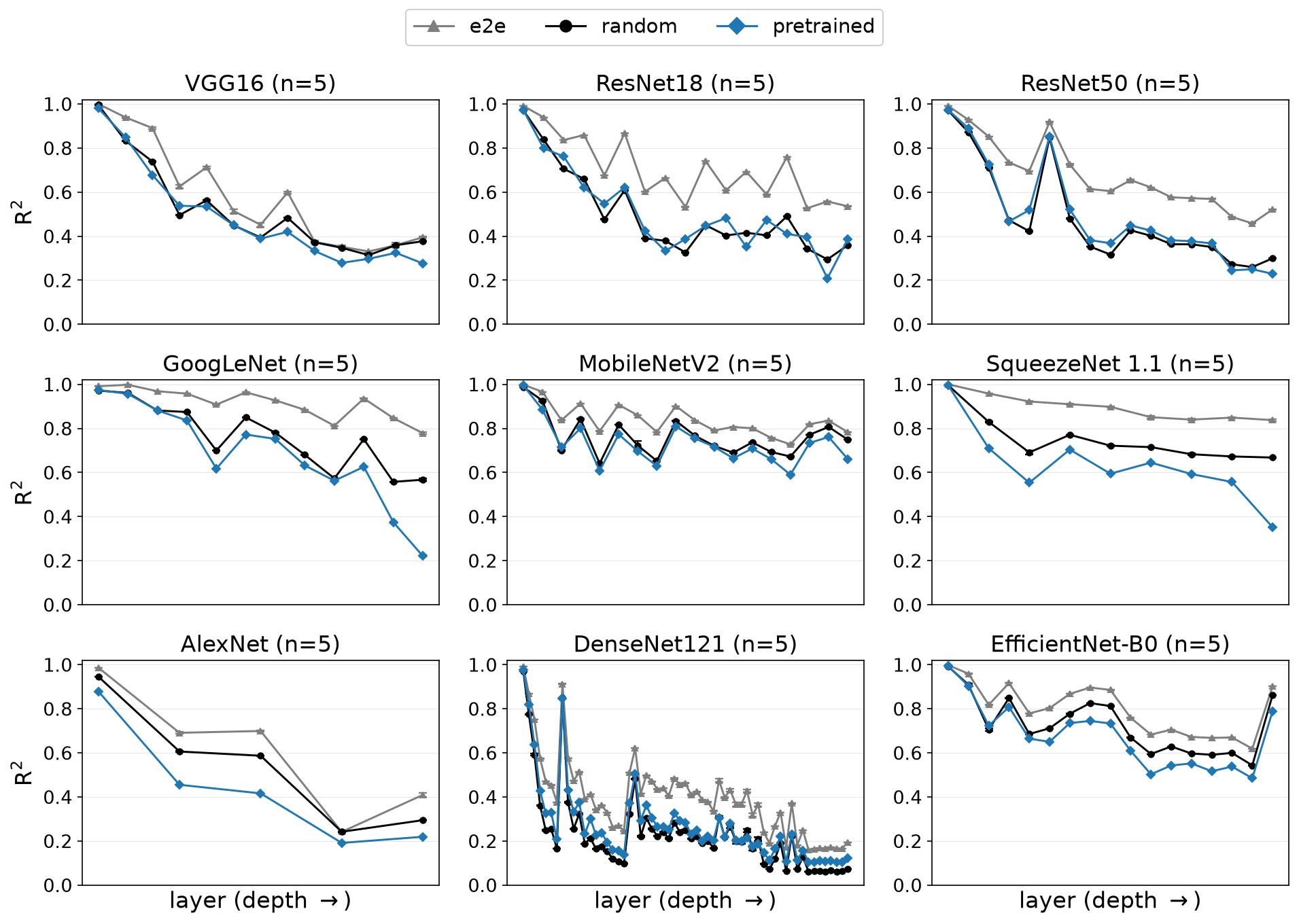}
   \caption{Recoverability ($R^2$) over depth for convolutional layers with kernel size $K>1$ across 9 CNN architectures, comparing pretrained, randomly-initialized, and end-to-end (e2e) trained encoders. \textbf{Top row:} VGG16, ResNet18, ResNet50. \textbf{Middle row:} GoogLeNet, MobileNetV2, SqueezeNet 1.1. \textbf{Bottom row:} AlexNet, DenseNet121, and EfficientNet-B0. Recoverability tends to decrease over depth for most models, with the exception of MobileNet and EfficientNet.}
  \label{fig:architecture_comparison}
 \end{figure}

We find that recoverability generally decreases over layer depth across architectures. Figure~\ref{fig:architecture_comparison} shows recoverability for the three encoders (pretrained, random, and e2e) for layers with kernel size $K>1$. We plot only these layer types because we find a consistent trend between recoverability and layer type. For example, a ResNet50 block consists of a $1\times1$ channel reduction, a $3\times3$ convolution, and a $1\times1$ channel expansion. We see that $R^2_{pre} \approx 1.0$ for $1\times1$ expand layers at all depths, whereas for $1\times1$ reduce layers, $R^2_{pre}$ starts at 0.84 and decreases over depth. Figure~\ref{fig:layer_type_comparison} in Appendix~\ref{app:layer_gap} illustrates this trend, where we plot pretrained recoverability for all layers in each model, colored by layer type. 

Among the layers with $K>1$, pretrained recoverability falls from about 1.0 in the first layer to between 0.1 and 0.4 in the last layer for VGG16, ResNet18, ResNet50, GoogLeNet, SqueezeNet, AlexNet, and DenseNet121. MobileNetV2 exhibits a less pronounced decrease in recoverability over depth (from 0.99 to 0.66), and EfficientNet-B0's last layer recoverability rises to 0.79 from the previous layer's 0.49. The e2e encoder, which is trained to reconstruct the patches, lies above both the pretrained and the random encoder in nearly every layer, so its distance from them indicates how much a better encoder of the same width could recover.

\subsection{A layer's recoverability is strongly correlated with its effective compression}
\begin{figure}[!ht]
  \centering
  \includegraphics[width=0.85\linewidth]{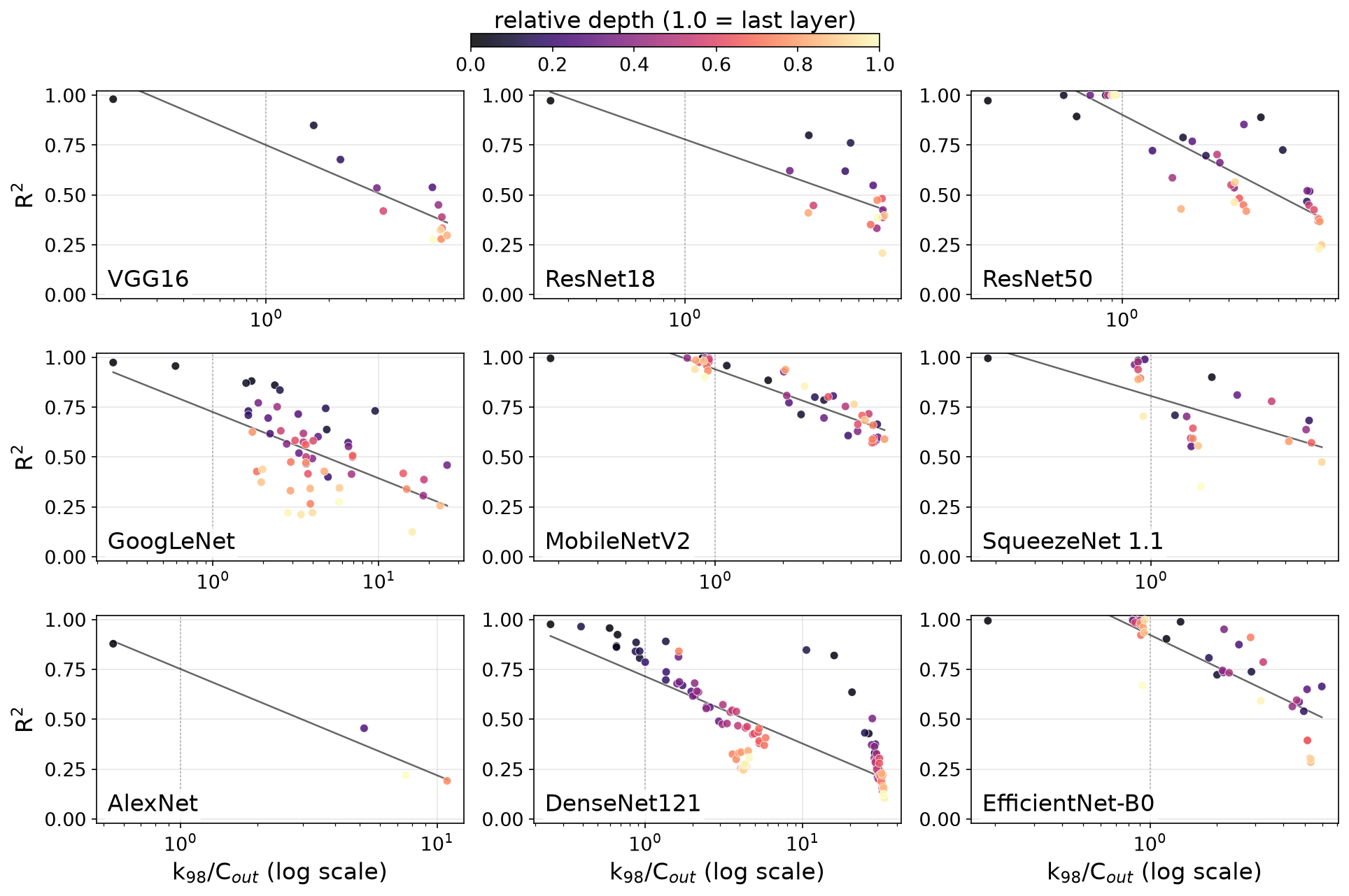}
   \caption{Pretrained recoverability ($R^2_{pre}$) against effective compression $k_{98}/C_{out}$ (log scale) for each of the nine architectures. Each point is one layer, colored by relative depth (0 = first layer, 1.0 = last layer). The black line is a least-squares fit of $R^2_{pre}$ on $\log(k_{98}/C_{out})$ and the dotted vertical line marks $k_{98}=C_{out}$, where a layer stops being a bottleneck. Recoverability is strongly correlated with effective compression in seven of the nine models ($\rho=-0.61$ to $-0.91$) and moderately so in GoogLeNet ($\rho=-0.55$); AlexNet has too few layers to analyze.}
  \label{fig:recoverability_vs_effective_compression}
 \end{figure}

Layer types have different magnitudes of recoverability, which follows from their compression. A layer's type is set by $C_{in}$, $C_{out}$, and $K$, and these also determine its nominal compression ratio $d/C_{out}=C_{in}\cdot K^2/C_{out}$. A $1\times1$ expansion never has a bottleneck, a $1\times1$ reduction always has a nominal one, and a layer with $K>1$ has one only when $C_{in}K^2>C_{out}$. Nominal compression can mislead, however, because input patches may occupy far fewer than $d$ directions. We therefore use the effective compression, $k_{98}/C_{out}$, where $k_{98}$ is the number of principal components that explain 98\% of the variance in a layer's input patches. Figure~\ref{fig:recoverability_vs_effective_compression} plots recoverability over effective compression. Recoverability falls as effective compression rises in every architecture, and layers with effective compression below 1 sit near $R^2\approx1$ no matter their depth: the $1\times1$ expand layers of ResNet50, MobileNetV2, SqueezeNet, and EfficientNet-B0 have an effective compression of 0.88 to 0.94.

Effective compression is strongly correlated with recoverability, with Spearman $\rho$ ranging from $-0.55\text{ to } -0.91$ across models, and it is a stronger predictor than nominal compression in all models as well; nominal compression is not significant in ResNet18 ($\rho=-0.34$, $p=.19$). Depth is a strong predictor in networks with a single repeated layer type (VGG16, ResNet18, GoogLeNet: $\rho=-0.97$, $-0.79$, $-0.81$) but weaker where layer types alternate (ResNet50, MobileNetV2: $\rho=-0.28$, $-0.18$, not significant).

The anomalies we find in recoverability over depth are well explained by effective compression. Four ResNet50 layers in stage 2 (\texttt{layer2\_0} to \texttt{layer2\_3}, second convolution) have the same patch dimension ($d=1152$), width ($C_{out}=128$), and nominal compression ($9\times$), yet their recoverability is 0.52, 0.85, 0.52, and 0.38. Their effective compression is $6.9\times$, $3.5\times$, $6.8\times$, and $7.6\times$, so the spike in recoverability coincides with a dip in effective compression.

\subsection{Trends in learning-attributed recoverability differ across models}
\begin{figure}[H]
  \centering
  \includegraphics[width=0.55\linewidth]{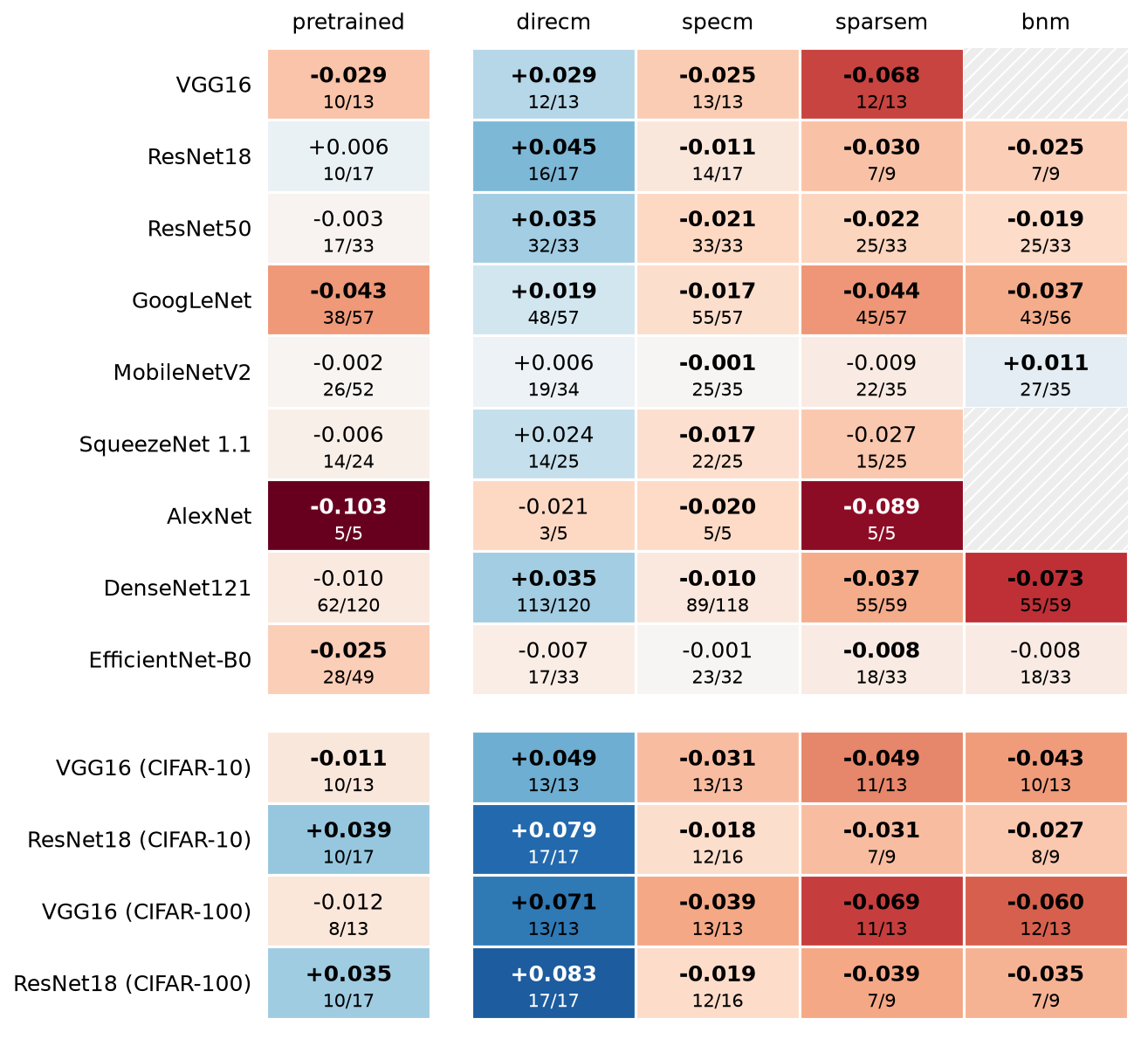}
   \caption{Effect of transferring one pretrained property onto a random encoder, averaged over layers. Rows are the nine ImageNet models followed by VGG16 and ResNet18 trained on CIFAR-10 and CIFAR-100. The left column is the net effect of training, mean $R^2_{pre}-R^2_{rand}$ over layers; the other columns are the effect of matching the pretrained singular vectors (\texttt{direcm}), singular-value spectrum (\texttt{specm}), activation sparsity (\texttt{sparsem}), or BatchNorm scale and shift (\texttt{bnm}), mean $R^2_{control}-R^2_{rand}$ over layers. Blue cells represent a positive mean difference, and red cells represent a negative mean difference. The fraction under each value is the number of layers with the same sign as the mean; bold marks a 95\% bootstrap CI over layers that excludes zero. Each column except for pretrained has a mostly consistent direction of change from the random initialization across models, positive for \texttt{direcm}, and negative for \texttt{specm}, \texttt{sparsem}, and \texttt{bnm}.}
  \label{fig:property_effects_heatmap.png}
 \end{figure}

Thus far, we have found that recoverability depends on effective compression and tends to decrease over depth. Figure~\ref{fig:property_effects_heatmap.png}'s left column shows that learning-attributed recoverability, $R^2_{pre}-R^2_{rand}$, is not uniform across architectures: its mean over layers ranges from $-0.103$ in AlexNet to $+0.039$ in ResNet18 trained on CIFAR-10 and is negative in 10 of 13 models. A negative value means that the pretrained weights recover less of the input than random weights. Learning-attributed recoverability also differs by layer type within a model. In SqueezeNet, $1\times1$ reduce layers have positive learning-attributed recoverability (mean $+0.167$) while spatial convolutions are negative ($-0.130$); in DenseNet121, spatial convolutions are positive ($+0.037$) and $1\times1$ reduce layers negative ($-0.059$) (Figure~\ref{fig:architecture_comparison_gap}, Appendix~\ref{app:layer_gap}). We do not find a trend in learning-attributed recoverability that is common to all models.

\subsection{Copying a single property of pretrained weights to the random initialization moves recoverability in the same direction across models}
To ask which properties of the pretrained weights produce this gap, we transfer one property at a time onto a random encoder and measure its effect, mean $R^2_{control}-R^2_{rand}$ over layers (Figure~\ref{fig:property_effects_heatmap.png}). Each control changes recoverability with a consistent sign across models. Matching the pretrained activation sparsity (\texttt{sparsem}) lowers recoverability in all 13 models, as does the pretrained spectrum (\texttt{specm}), and the pretrained BatchNorm parameters (\texttt{bnm}) do so in 9 of 10 models with BatchNorm. Transferring the pretrained singular vectors (\texttt{direcm}) raises recoverability in 11 of 13 models. Because these effects have opposite signs, they can offset each other: in ResNet50, the net gap is $-0.003$ while the effects of singular vectors, spectrum, and activation sparsity are $+0.035$, $-0.021$, and $-0.022$. Each control isolates one property, so the effects indicate the direction and size of that property's effect. We encourage future work to develop more random controls that account for different properties of pretrained weights.

Since each control moves recoverability in a consistent direction, we ask whether any single control explains the learning-attributed gap. Recall that a control explains the gap of a model or layer type if it reduces the summed squared gap over layers by at least half (Equation~\ref{eq:explains}). We find that no single control explains the gap across models. However, different controls explain the gap in five of the nine ImageNet models: activation-sparsity matching in AlexNet and SqueezeNet, BatchNorm transfer in DenseNet121 and EfficientNet-B0, and the pretrained singular vectors in MobileNetV2 (Appendix~\ref{app:controls}). We find that in SqueezeNet (Figure~\ref{fig:gap_by_type_cases_squeezenet}, Appendix~\ref{app:layer_gap}), \texttt{direcm} explains essentially all of the gap in the $1\times1$ reduce layers, while \texttt{sparsem} explains 89\% and 95\% of the gap in the $K>1$ convolution and $1\times1$ expand layers, including the sharp drop in the deepest layers. Other layer types are explained by no control: in GoogLeNet's spatial convolutions (mean gap $-0.10$), the best control, \texttt{specm}, explains 32\%.

\subsection{Deep layers of the same network trained on different classification tasks exhibit different recoverability profiles}

\begin{figure}[H]
  \centering
  \includegraphics[width=0.80\linewidth]{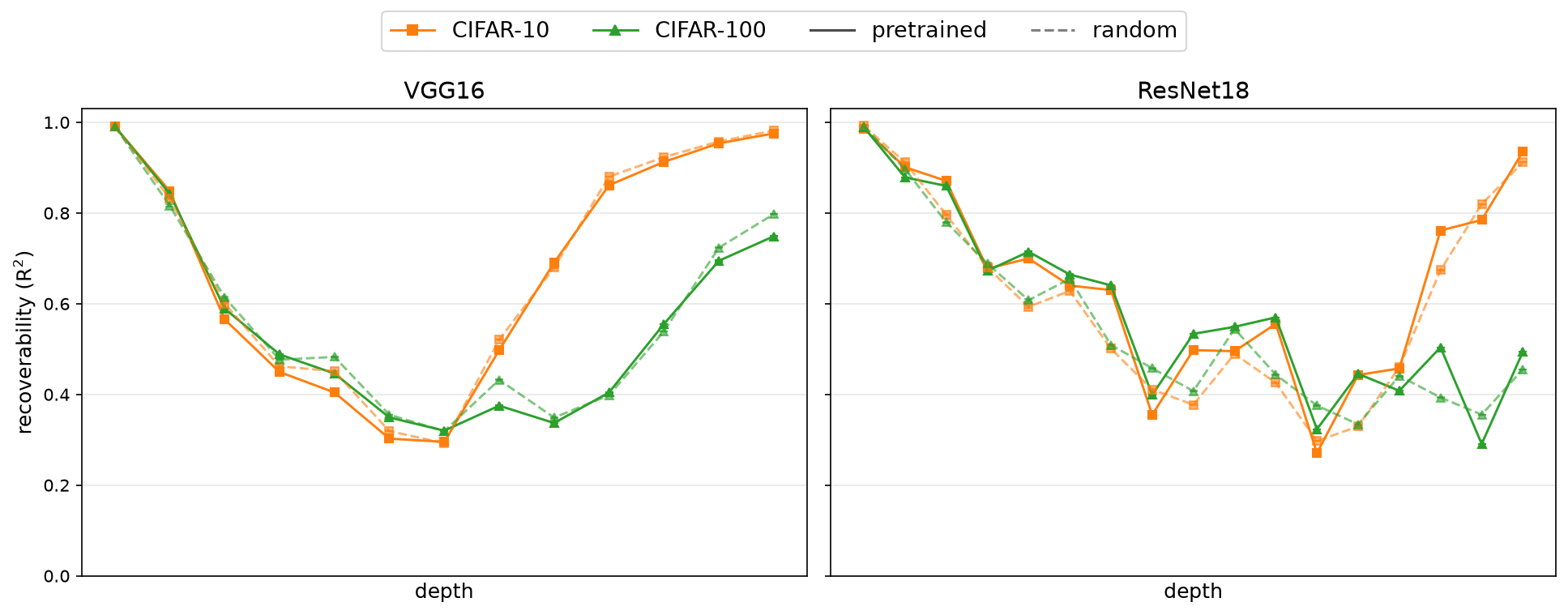}
   \caption{Recoverability ($R^2$) over depth for VGG16 (left) and ResNet18 (right) trained on CIFAR-10 and CIFAR-100, for pretrained (solid) and random (dashed) encoders. Error bars are the standard error over probe seeds (5 for CIFAR-10, 3 for CIFAR-100) and are smaller than the markers. Deep layer recoverability increases more for the CIFAR-10 trained variants than the CIFAR-100 trained variants.}
  \label{fig:imagenet_vs_cifar_cifar100_combined}
 \end{figure}

We next assess recoverability over depth for VGG16 and ResNet18 trained on CIFAR-10 versus CIFAR-100, as shown in Figure~\ref{fig:imagenet_vs_cifar_cifar100_combined}. We compare the two CIFAR datasets directly because the CIFAR versions of these networks differ in structure from the ImageNet ones (native $32\times32$ input, a CIFAR stem for ResNet18, and BatchNorm in VGG16). In both models, recoverability for CIFAR-10 tracks CIFAR-100 through the early and middle layers and then rises more in the deep layers, to 0.98 versus 0.75 in the last layer of VGG16 and 0.93 versus 0.49 in ResNet18. We also see that the pretrained and random traces track each other closely across depth. The mean difference between CIFAR-100 and CIFAR-10 is the same for the two encoders (VGG16: $-0.124$ pretrained and $-0.123$ random; ResNet18: $-0.060$ and $-0.057$), which suggests that this difference comes from the layer inputs and not from how the trained encoder treats them.

\section{Conclusion}
We introduced layer-wise input recoverability: the fraction of variance in the patches a convolutional layer reads that a decoder can reconstruct from the layer's output. Probing each layer on its own input patches and their representations, rather than its entire input feature map and the original image, attributes recoverability to a single operation and ties the probe's size to channels and kernel size, which lets us apply our method to nine modern deep CNNs.
We also introduced learning-attributed recoverability that compares pretrained recoverability with a random baseline, and we find it is not uniform across architectures. To ask what produces it, we introduced single-property random controls that transfer one pretrained property onto a random encoder. Each control changes recoverability with a consistent sign across models: learned singular vectors raise it, while the learned spectrum and activation sparsity lower it. We find that some models' learning-attributed recoverability can be explained by a single control, but there is no control that explains the phenomenon across models. Finally, effective compression, the number of principal components that hold 98\% of a layer's input variance relative to its width, was the most consistent predictor of recoverability across architectures, and it explains anomalies that depth and nominal compression do not.

Our work has limitations. Recoverability is decoder-relative: in layers with a real bottleneck, wider decoders keep improving it, so our values are lower bounds on what a stronger decoder could recover. Our random controls change one property at a time, so they show how recoverability is associated with a property and not that the property causes it, and they do not cover interactions between properties or every property of pretrained weights.

Because patch-level probing scales to deep CNNs, we encourage future work to use Shannon information estimators at the patch level, which would align our analysis more closely with the information bottleneck framework.

\section*{AI use statement}
Following the ICLR AI Policy for Authors, we disclose the following. Among the tasks for which
disclosure is \emph{required}, generative AI was used to \textbf{implement methods} --- writing and
refactoring the experiment, measurement and figure-generation scripts in the accompanying repository ---
and to \textbf{reformat data}. It was also used to \textbf{interpret results} in
the limited sense of drafting candidate readings of measured numbers, each of which the authors checked
against the underlying artifacts before adoption. We did \emph{not} use generative AI to generate
synthetic datasets, develop theoretical models or conceptual frameworks, formulate mathematical claims,
prove or assist in proving them, propose or refine the hypotheses, or design the experiments; the
remaining required-disclosure categories (translation, qualitative and thematic analysis) are not
applicable to this work. Among the tasks for which disclosure is \emph{recommended}, we used it to
create and edit software code, create and modify figures, draft and edit parts of this paper for
readability, summarize and identify related literature, and format references.

We have reviewed all AI-assisted work. Every number reported here is produced by code in the repository
and re-derived by the automated checkers described above rather than generated by a model; the checkers
exist precisely so that no quoted value depends on a model's recollection, and they guard derived
quantities (ratios, shares and the direction of comparisons) as well as directly measured ones.
AI-drafted prose was verified claim-by-claim against those artifacts, and literature identified with AI
assistance was confirmed against the primary sources before citation. All claims and their scoping are
the authors', and we take full responsibility for the final content of this work, including text, claims
and artifacts produced with the aid of generative AI.

\clearpage
\bibliography{iclr2027_conference}
\bibliographystyle{iclr2027_conference}

\appendix

\section{Appendix}
\subsection{Patch sampling ablations}
\label{app:patches}

Our main experiments draw one patch from each of 40,000 training and 8,000 validation images at every probed layer (Section~\ref{sec:patches}). We test whether recoverability depends on how this patch budget is spent, using ResNet18 trained on CIFAR-10, five layers spanning depth (the stem, \texttt{layer1\_1\_conv2}, \texttt{layer2\_0\_conv1}, \texttt{layer3\_1\_conv1}, \texttt{layer4\_1\_conv2}), pretrained and random encoders, the linear decoder, and three patch seeds. Training and validation images are disjoint in every configuration, so more patches per image cannot leak information between splits.

\paragraph{Fixed total number of patches.} We keep 50,000 training and 10,000 validation patches per layer and spread them over fewer images: 50,000 images with 1 patch each, 25,000 with 2, 12,500 with 4, and 3,125 with 16 (validation images scale in the same proportion). Compared with the main setting, the largest change in pretrained $R^2$ over layers and configurations is 0.0034 and the largest change in learning-attributed recoverability $R^2_{pre}-R^2_{rand}$ is 0.0018. The sign of learning-attributed recoverability never changes.

\begin{figure}[h]
 \centering
 \includegraphics[width=1\linewidth]{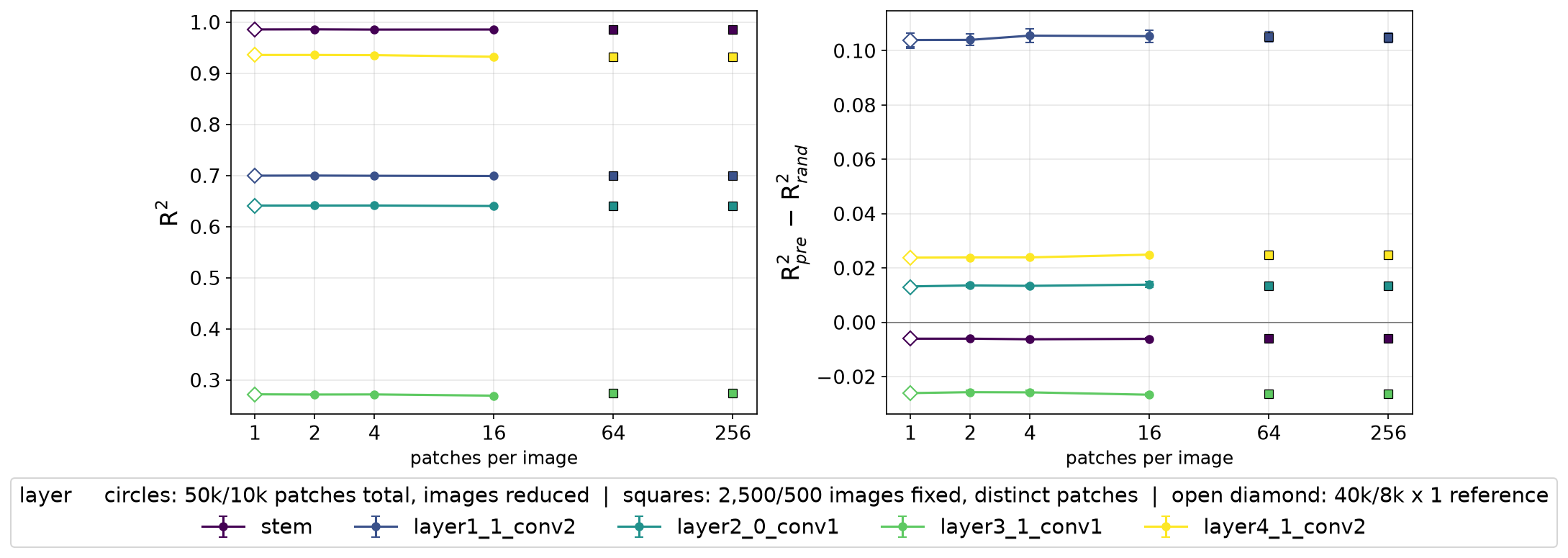}
 \caption{Patches-per-image ablation for ResNet18 trained on CIFAR-10. \textbf{Left:} pretrained recoverability ($R^2$). \textbf{Right:} learning-attributed recoverability ($R^2_{pre}-R^2_{rand}$). One line per layer; circles keep 50,000/10,000 patches in total and reduce the number of images, squares fix 2,500/500 images and sample 64 or 256 distinct patches per image, and the open diamond is the main-text setting (40,000/8,000 images, one patch each). Error bars are standard errors over three patch seeds.}
 \label{fig:app_patch_cifar}
\end{figure}

\paragraph{Fixed number of images.} We then fix 2,500 training and 500 validation images and sample 64 or 256 distinct patches per image without replacement (all valid positions where a layer has fewer: 36 in \texttt{layer3\_1\_conv1} and 4 in \texttt{layer4\_1\_conv2}). Recoverability changes by at most 0.0034 and learning-attributed recoverability by at most 0.0015, again with no change of sign (squares in Figure~\ref{fig:app_patch_cifar}).

\begin{figure}[h]
 \centering
 \includegraphics[width=1\linewidth]{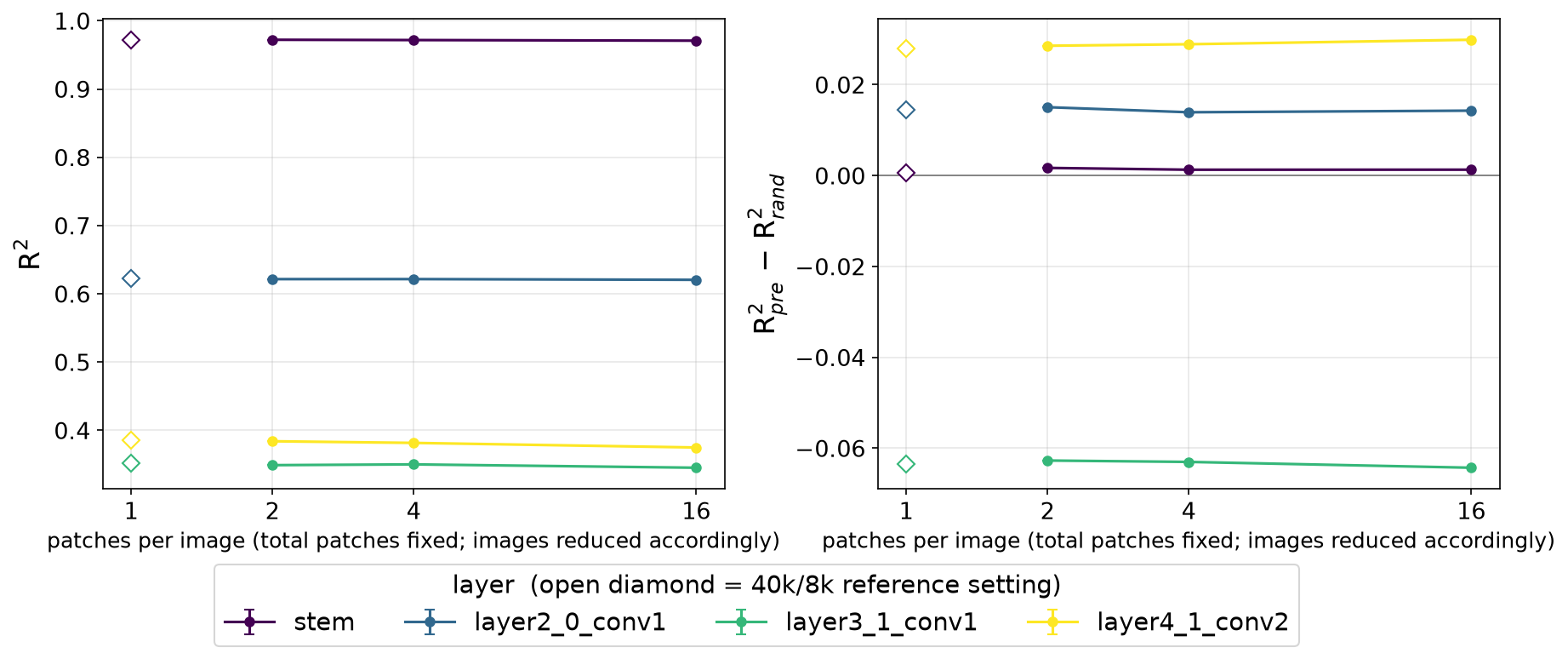}
 \caption{Patches-per-image ablation for ResNet18 trained on ImageNet, with a fixed total of 40,000/8,000 patches (one patch seed).}
 \label{fig:app_patch_imagenet}
\end{figure}

\paragraph{ImageNet.} For ImageNet-trained ResNet18 (four layers, one patch seed), the ImageNet validation set has only 50,000 images, so we keep the total at 40,000/8,000 patches and use 2, 4, and 16 patches per image (20,000, 10,000, and 2,500 training images). The largest change in pretrained $R^2$ is 0.0104 and in learning-attributed recoverability 0.0019, with no change of sign (Figure~\ref{fig:app_patch_imagenet}).

Together, these results show that recoverability and learning-attributed recoverability do not depend on whether patches come from many images or are densely sampled from few, consistent with neighboring patches being highly redundant.

\subsection{Decoder capacity, saturation, and memorization}
\label{app:capacity}

Recoverability is decoder-relative: a more expressive decoder can recover more, and a decoder large enough to memorize its training set can also fit noise. Our primary quantity uses the linear decoder, which is the exact least-squares solution given the encoder and involves no optimizer, initialization, or early stopping. To check how much this choice matters, we evaluate a ladder of linear-plus-residual-MLP decoders of hidden width $m \cdot C_{out}$, $m\in\{1,4,16\}$, on four layers each of AlexNet, ResNet18, ResNet50, SqueezeNet 1.1, VGG16, and for the three encoders (pretrained, random, and e2e). For every layer, we report two diagnostics. The \emph{saturation gap} is $|R^2_{16\times}-R^2_{4\times}|$, which is small when a wider decoder no longer helps. The \emph{memorization control} is the held-out $R^2$ of the same decoder class fit to shuffled targets, which measures how much the decoder can fit without structure in the data. Assessment of memorization is a common technique for evaluating probes.

\begin{figure}[h]
 \centering
 \includegraphics[width=1\linewidth]{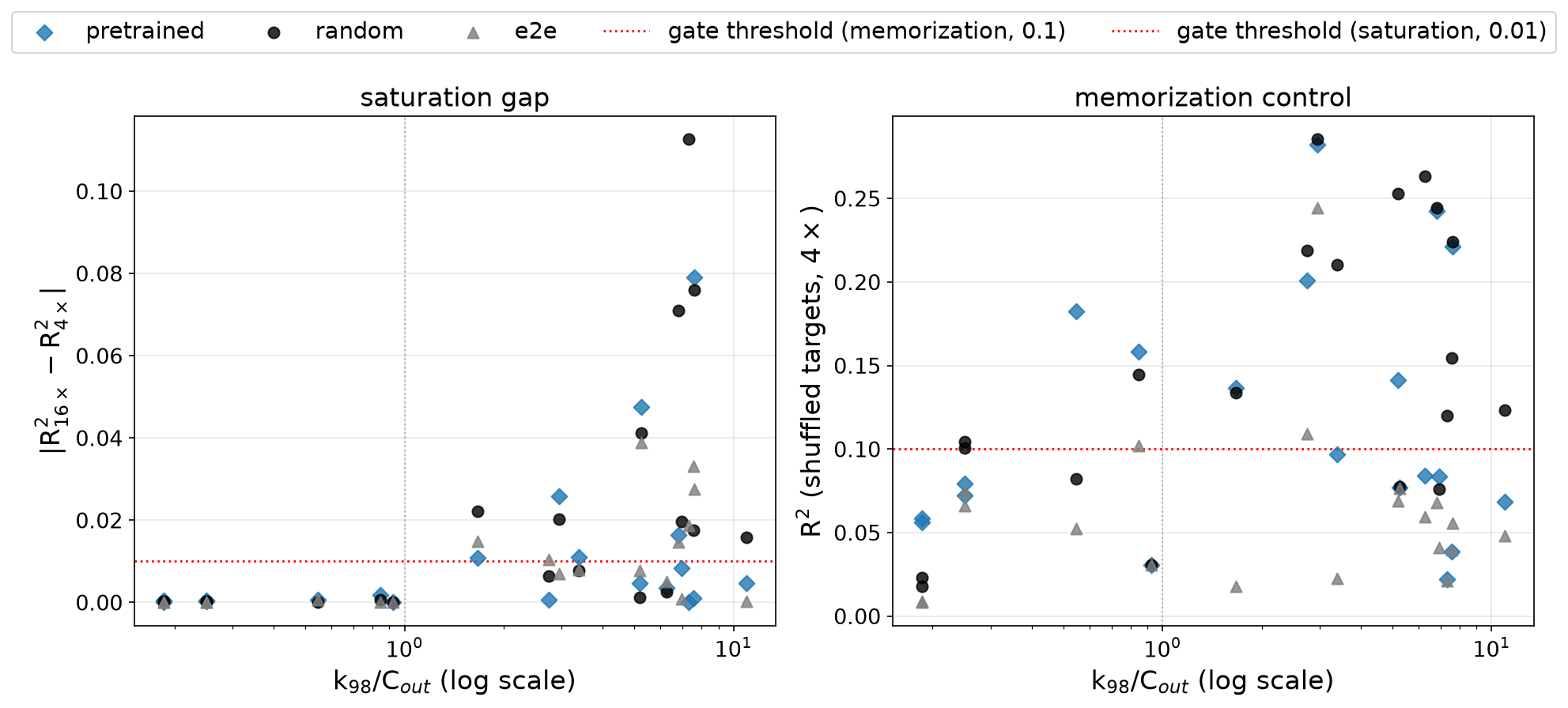}
 \caption{Decoder saturation and memorization against effective compression $k_{98}/C_{out}$, one point per (architecture, layer, encoder). \textbf{Left:} saturation gap $|R^2_{16\times}-R^2_{4\times}|$ against the gate threshold of 0.01. \textbf{Right:} memorization-control $R^2$ at $4\times$ against the gate threshold of 0.1. Failures of the saturation gate lie at $k_{98}/C_{out}>1$.}
 \label{fig:app_capacity_compression}
\end{figure}

\begin{figure}[h]
 \centering
 \includegraphics[width=1\linewidth]{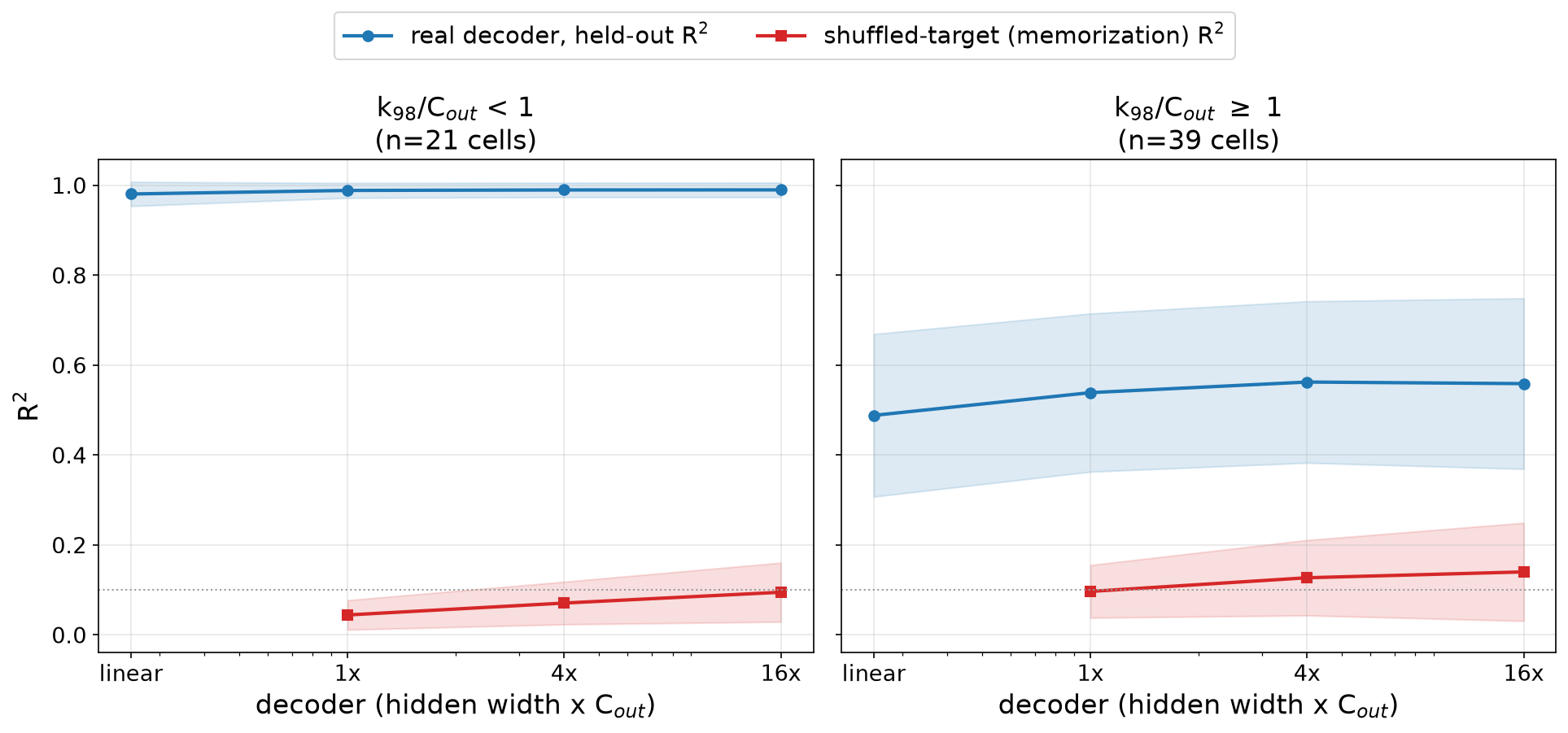}
 \caption{Held-out $R^2$ of the real decoder and of the shuffled-target memorization control against decoder width ($C_{out}$ multiples), for layers with $k_{98}/C_{out}<1$ (left, $n=21$) and $k_{98}/C_{out}\ge1$ (right, $n=39$). Lines are means and shading is $\pm1$ standard deviation over cells; the dotted line is the memorization gate at 0.1.}
 \label{fig:app_capacity_summary}
\end{figure}

We set two decoder-quality gates before running the study: at least 80\% of layers with a saturation gap below 0.01, and at least 90\% of layers with a memorization $R^2$ below 0.1. Both fail (63\% and 58\% of the 60 layers; mean memorization $R^2$ is 0.107 at $4\times$ and 0.124 at $16\times$). The failures are concentrated in layers that are real bottlenecks (Figure~\ref{fig:app_capacity_compression}): of the 21 cells with effective compression $k_{98}/C_{out}<1$, all are saturated (median gap 0.000), whereas only 44\% of the 39 cells with $k_{98}/C_{out}\ge1$ are, and the saturation gap correlates with effective compression (Spearman $\rho=+0.63$). Memorization rises with decoder width in compressive layers (Figure~\ref{fig:app_capacity_summary}), but correlates only weakly with effective compression ($\rho=+0.18$). Thus, for layers with a real bottleneck, recoverability under a fixed decoder is a lower bound that wider decoders keep tightening, and we do not claim it is saturated.

What our comparisons need is not saturation but stability of differences between encoders. Between the linear decoder and the $16\times$ decoder, the sign of the pretrained-minus-random difference is unchanged in 18 of 20 layer-model pairs, and the median absolute change in this difference is 0.010 (maximum 0.061). We therefore report the linear decoder in the main text, and treat effects smaller than about 0.01 as within decoder sensitivity.

\begin{figure}[h]
 \centering
 \includegraphics[width=1\linewidth]{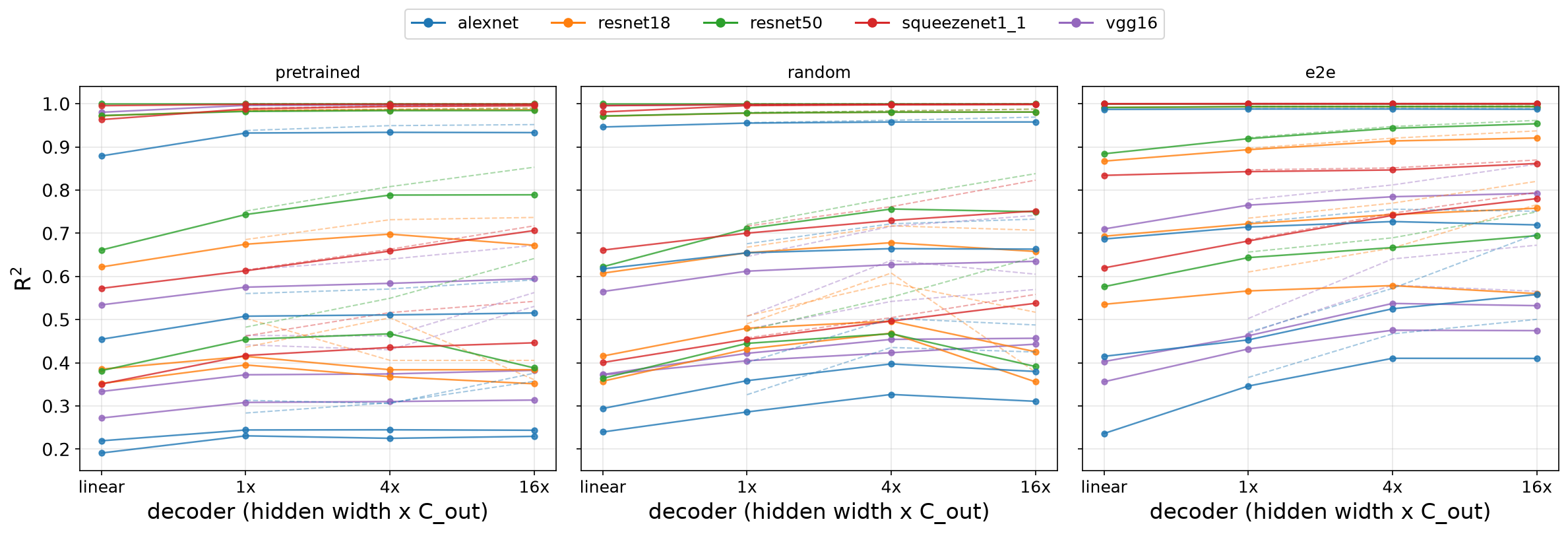}
 \caption{Held-out recoverability (solid) and training recoverability (dashed) against decoder size for each (architecture, layer) in the capacity study, for pretrained, random, and e2e encoders.}
 \label{fig:app_capacity_dose}
\end{figure}

\subsection{Simple examples with directly computable information}
\label{app:simple}

Recoverability is a decoder-relative quantity, so it is a V-information-style measure and not an estimate of Shannon information. The purpose of this appendix is to check how the two relate in simple settings where the information is directly computable, so that no estimator is needed. In each setting below, we compute the information exactly and compare it with recoverability.

\subsubsection{Binary image patches}
Let $X$ be a $3\times3$ binary patch of a natural image (CIFAR-10 images, each binarized at its own mean), which takes at most 512 values with an empirical distribution known exactly. Encoders $z=f(Wx+b)$ have width $C\in\{2,3,4,6,8,12\}$, activation ReLU, $\tanh$, or identity, and $W$ that is random, orthogonal, low-rank, trained end-to-end to reconstruct $X$, or trained for a task label (a stand-in for a pretrained network). A deterministic real-valued encoder of a discrete input is injective, so $H(X\mid Z)=0$ and carries no information about the encoder; we therefore observe $z$ through Gaussian noise of relative standard deviation $\rho\in\{0.1,0.3,1.0\}$. For every encoder we compute $H(X\mid Z)$ and the Bayes-optimal decoder without estimation, the exact optimal linear decoder, and a trained linear-plus-residual-MLP decoder of the same class as in the main experiments.

\begin{figure}[h]
 \centering
 \includegraphics[width=1\linewidth]{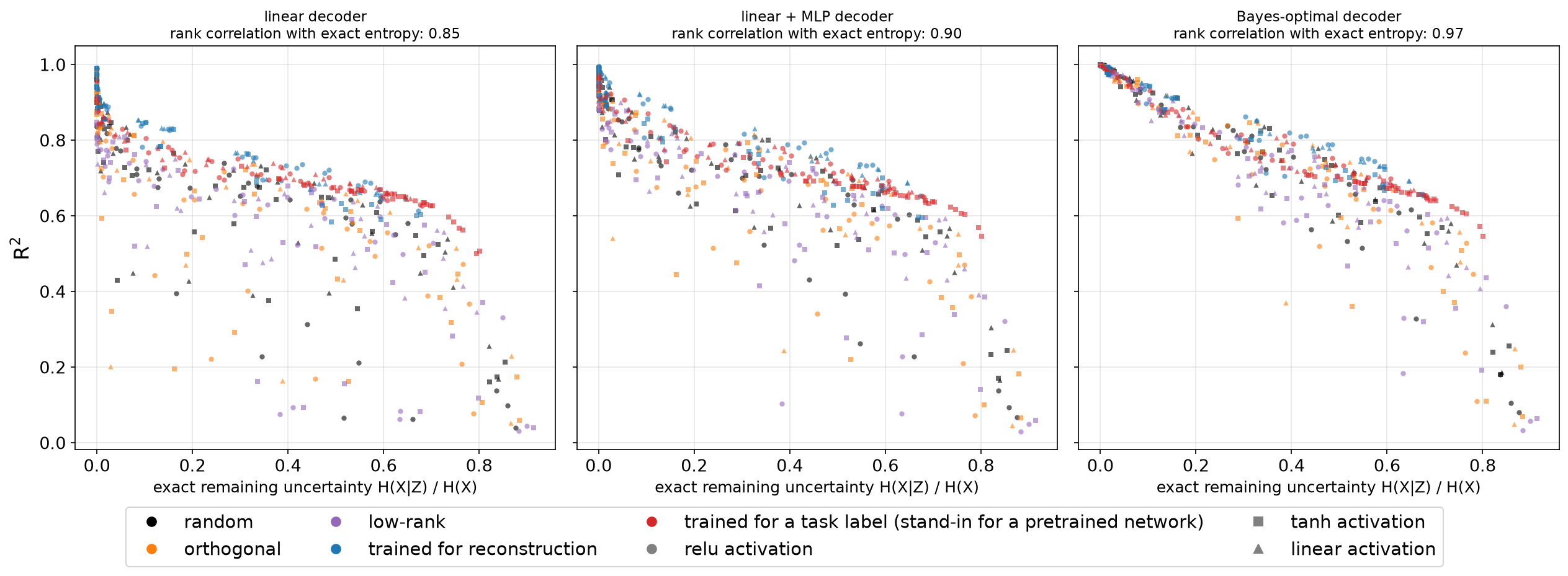}
 \caption{Recoverability ($R^2$) against exact remaining uncertainty $H(X\mid Z)/H(X)$ for $3\times3$ binary image patches, for a linear decoder (left), a linear-plus-residual-MLP decoder (middle), and the Bayes-optimal decoder (right). Each point is one encoder; color gives the encoder type and marker the activation function.}
 \label{fig:app_simple_discrete}
\end{figure}

Figure~\ref{fig:app_simple_discrete} plots recoverability against the exact remaining uncertainty $H(X\mid Z)/H(X)$. The rank correlation between recoverability and the information retained, $1-H(X\mid Z)/H(X)$, is 0.85 for the linear decoder, 0.90 for the linear-plus-MLP decoder, and 0.97 for the Bayes-optimal decoder, so recoverability tracks information when the decoder is optimal and less closely when it is restricted, as V-information predicts. Among pairs of encoders with the same width, activation, and noise, and an information difference above 0.1 bit, the sign of the recoverability difference agrees with the sign of the information difference for reconstruction-trained versus random encoders in 98.6\%, 97.2\%, and 98.6\% of pairs (linear, MLP, Bayes decoder), for low-rank versus random in 85.4\%, 92.4\%, and 96.2\%, for orthogonal versus random in 75.0\%, 83.6\%, and 91.4\%, and for task-trained versus random in 67.8\%, 76.6\%, and 90.1\%. Disagreement therefore concentrates where encoders keep different things, and is largest for the comparison closest to pretrained-versus-random, task-trained versus random, where a linear decoder is least reliable.

\subsubsection{Gaussian patch models}
We next take $X\sim\mathcal N(0,\Sigma)$ with $\Sigma$ as the patch covariance of four ResNet18 layers (the stem, \texttt{layer1\_0\_conv1}, \texttt{layer2\_0\_conv1}, \texttt{layer3\_0\_conv1}), computed on CIFAR-10 train images passed through the pretrained network. Encoders are linear: the layer's pretrained filters, a default random draw, an orthogonal draw, and random directions with the pretrained spectrum, each scaled to equal output power and observed through noise of variance $\gamma\in\{10^{-3},10^{-2},10^{-1},1\}$ relative to output power. Mutual information $I(X;Z)=\tfrac12\log\det(I+W\Sigma W^{\top}/\gamma)$ and the optimal linear reconstruction error are exact.

\begin{figure}[h]
 \centering
 \includegraphics[width=1\linewidth]{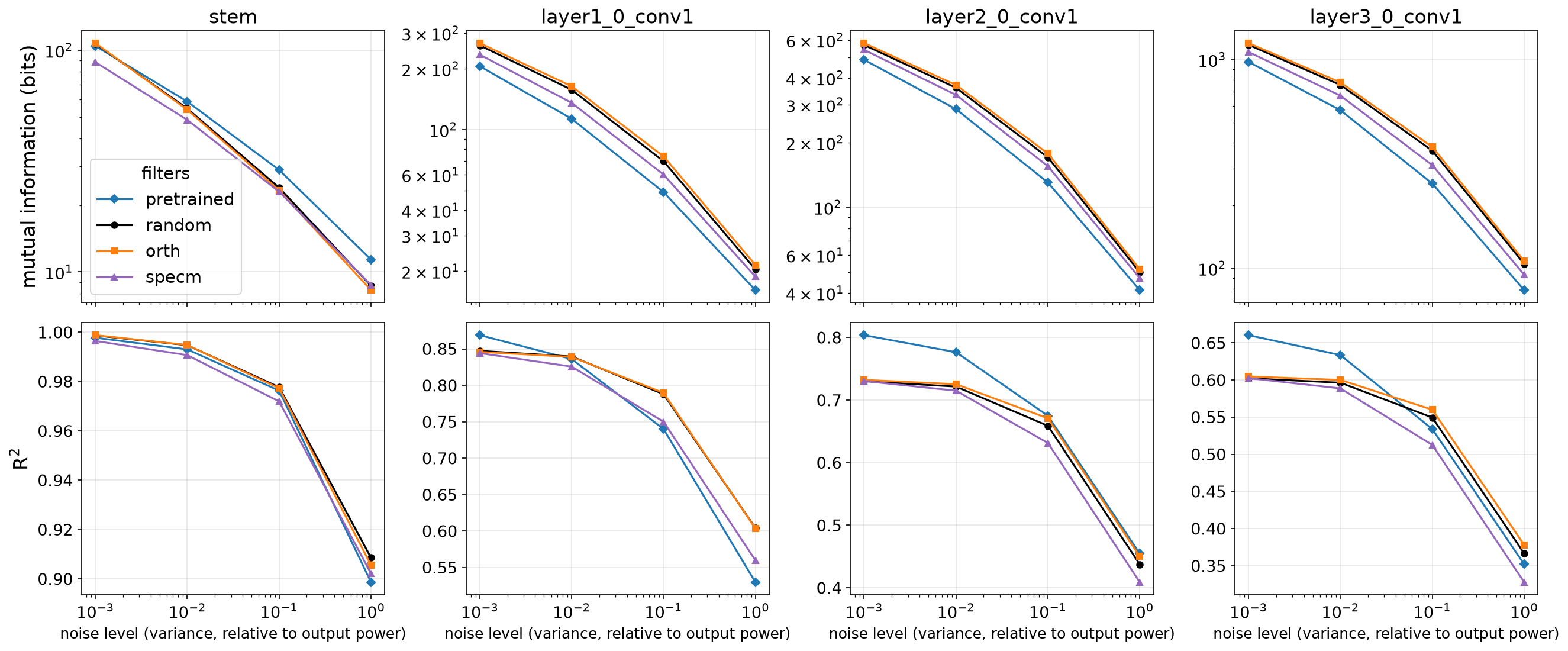}
 \caption{Gaussian patch models with covariances from four ResNet18 layers. \textbf{Top:} exact mutual information between input patches and noisy encoder outputs. \textbf{Bottom:} recoverability ($R^2$) of the optimal linear decoder, against noise level, for the layer's pretrained filters, random filters, orthogonal filters, and spectrum-matched filters (\texttt{specm}).}
 \label{fig:app_simple_gaussian}
\end{figure}

Recoverability and mutual information can rank encoders oppositely (Figure~\ref{fig:app_simple_gaussian}). Recoverability is variance-weighted (trace-based), whereas mutual information counts every direction the encoder resolves, including the low-variance tail of $\Sigma$ (log-determinant based). In the three layers past the stem, the pretrained filters carry less mutual information than random filters at every noise level (for example 206 versus 261 bits at $\gamma=10^{-3}$ in \texttt{layer1\_0\_conv1}), yet in \texttt{layer2\_0\_conv1} and \texttt{layer3\_0\_conv1} they are more recoverable at low noise. Over comparisons with a mutual-information difference above 0.1 bit, the recoverability ordering is opposite to the information ordering in 6 of 16 for pretrained versus random, 5 of 16 for pretrained versus spectrum-matched, and 13 of 15 for spectrum-matched versus random. The counts for the pretrained comparisons do not change when the covariance is estimated from four times as many patches (spectrum-matched versus random becomes 14 of 16). Recoverability thus depends more on the input's high-variance directions than information does, which is one reason we call the measure recoverability and not information. The effective sample is four layers, because the noise levels within a layer move together.

\subsection{Random control details}
\label{app:controls}

\paragraph{Baseline draw.} The random encoder uses the framework's default initialization with the layer's BatchNorm re-estimated on training patches. Because a default draw is closer to invertible than it might appear, we checked that the choice of draw does not drive our results. A semi-orthogonal draw of the same scale (orthonormal rows when $C_{out}\le d$, columns otherwise) changes $R^2$ by 0.004 to 0.014 root-mean-square across the nine ImageNet models, with a mean shift of $+0.002$ to $+0.009$, and it does not change the sign of any model's mean learning-attributed recoverability. We therefore keep the default draw and do not treat the orthogonal draw as a separate control.

\paragraph{Scale and spectrum shape.} Rescaling \texttt{specm} to the Frobenius norm of the random draw (spectrum shape at random's scale) changes $R^2$ by at most 0.001 RMS in VGG16, AlexNet, and SqueezeNet, so the \texttt{specm} effect is a property of the spectrum's shape and not of its scale relative to the bias. Similarly, replacing the random spectrum in \texttt{direcm} with a flat one (the pretrained singular vectors with equal singular values, called \texttt{polar} below) gives per-layer effects that correlate at 0.92 to 0.99 with \texttt{direcm} across models.

\paragraph{Which control accounts for the gap.} For each model, we score every control $b$ by the share of the summed squared gap over layers it removes, that is, one minus the left-hand side of Equation~\ref{eq:explains} with $R^2_b$ in place of $R^2_{control}$, and say a control explains the gap under the rule of Equation~\ref{eq:explains}, which requires a share of at least 0.5 and a root-mean-square gap of at least 0.02. Beyond the four controls in the main text, the candidates are the orthogonal draw (\texttt{orth}) and three composites: \texttt{specsp} (spectrum and activation sparsity matched), \texttt{specbn} (spectrum and BatchNorm scale and shift matched), and \texttt{gainxfer}, the BatchNorm-free counterpart of \texttt{bnm} (random directions rescaled to the pretrained per-channel weight norms, with the pretrained bias); \texttt{polar} is scored in place of \texttt{direcm}. Table~\ref{tab:app_controls} gives all scores. Eight candidates are scored per model, so a control that passes should be read as consistent with, and not as proof of, a mechanism. A control accounts for the gap in five of nine ImageNet models: activation-sparsity matching in AlexNet (0.87) and SqueezeNet (0.79), BatchNorm transfer in DenseNet121 (0.87) and EfficientNet-B0 (0.67), and preserved singular vectors in MobileNetV2 (0.77). None does in VGG16, ResNet18, ResNet50, GoogLeNet, or the four CIFAR models. Within a layer type, single controls sometimes account for more (Figure~\ref{fig:gap_by_type_cases_squeezenet}), but these per-type scores are exploratory because the rule was fixed per model.

\begin{table}[h]
\centering
\caption{Share of the summed squared gap over layers removed by each control (rows: models). Bold marks the highest-scoring control that reaches 0.5 where the RMS gap is at least 0.02; \texttt{polar} keeps the pretrained singular vectors with a flat spectrum. Dashes mark controls that are undefined for that model.}
\label{tab:app_controls}
\small
\begin{tabular}{lrrrrrrrrr}
\hline
model & RMS gap & \texttt{orth} & \texttt{bnm} & \texttt{specm} & \texttt{sparsem} & \texttt{specsp} & \texttt{specbn} & \texttt{polar} & \texttt{gainxfer} \\
\hline
VGG16 & 0.047 & $-0.16$ & -- & $+0.36$ & $-0.39$ & $-1.48$ & -- & $-1.99$ & $-0.04$ \\
ResNet18 & 0.055 & $-0.08$ & $+0.31$ & $+0.17$ & $+0.15$ & $-0.07$ & $+0.13$ & $-0.59$ & -- \\
ResNet50 & 0.035 & $-0.31$ & $+0.15$ & $-0.49$ & $-0.02$ & $-1.20$ & $-1.00$ & $-2.59$ & -- \\
GoogLeNet & 0.102 & $-0.13$ & $+0.31$ & $+0.27$ & $+0.31$ & $+0.39$ & $+0.42$ & $-0.26$ & -- \\
MobileNetV2 & 0.034 & $-0.04$ & $-0.27$ & $-0.10$ & $-8.95$ & $+0.26$ & $+0.24$ & $\mathbf{+0.77}$ & -- \\
SqueezeNet 1.1 & 0.141 & $-0.04$ & -- & $+0.10$ & $\mathbf{+0.79}$ & -- & -- & $+0.69$ & -- \\
AlexNet & 0.114 & $-0.04$ & -- & $+0.33$ & $\mathbf{+0.87}$ & -- & -- & $+0.21$ & -- \\
DenseNet121 & 0.060 & $-0.17$ & $\mathbf{+0.87}$ & $+0.18$ & $+0.63$ & -- & -- & $-0.73$ & -- \\
EfficientNet-B0 & 0.057 & $-0.05$ & $\mathbf{+0.67}$ & $-0.06$ & $+0.51$ & -- & -- & $+0.30$ & -- \\
VGG16 (CIFAR-10) & 0.020 & $-0.48$ & $-4.76$ & $-0.78$ & $-6.80$ & $-13.32$ & $-10.86$ & $-14.79$ & -- \\
ResNet18 (CIFAR-10) & 0.073 & $-0.01$ & $-0.20$ & $-0.06$ & $-0.29$ & $-0.66$ & $-0.55$ & $+0.16$ & -- \\
VGG16 (CIFAR-100) & 0.027 & $-0.33$ & $-4.05$ & $-0.84$ & $-5.90$ & -- & -- & $-12.00$ & -- \\
ResNet18 (CIFAR-100) & 0.079 & $-0.03$ & $-0.11$ & $0.00$ & $-0.19$ & -- & -- & $-0.08$ & -- \\
\hline
\end{tabular}
\end{table}

\subsection{Recoverability and learning-attributed recoverability by layer type}
\label{app:layer_gap}

\begin{figure}[t]
  \centering
  \includegraphics[width=0.92\linewidth]{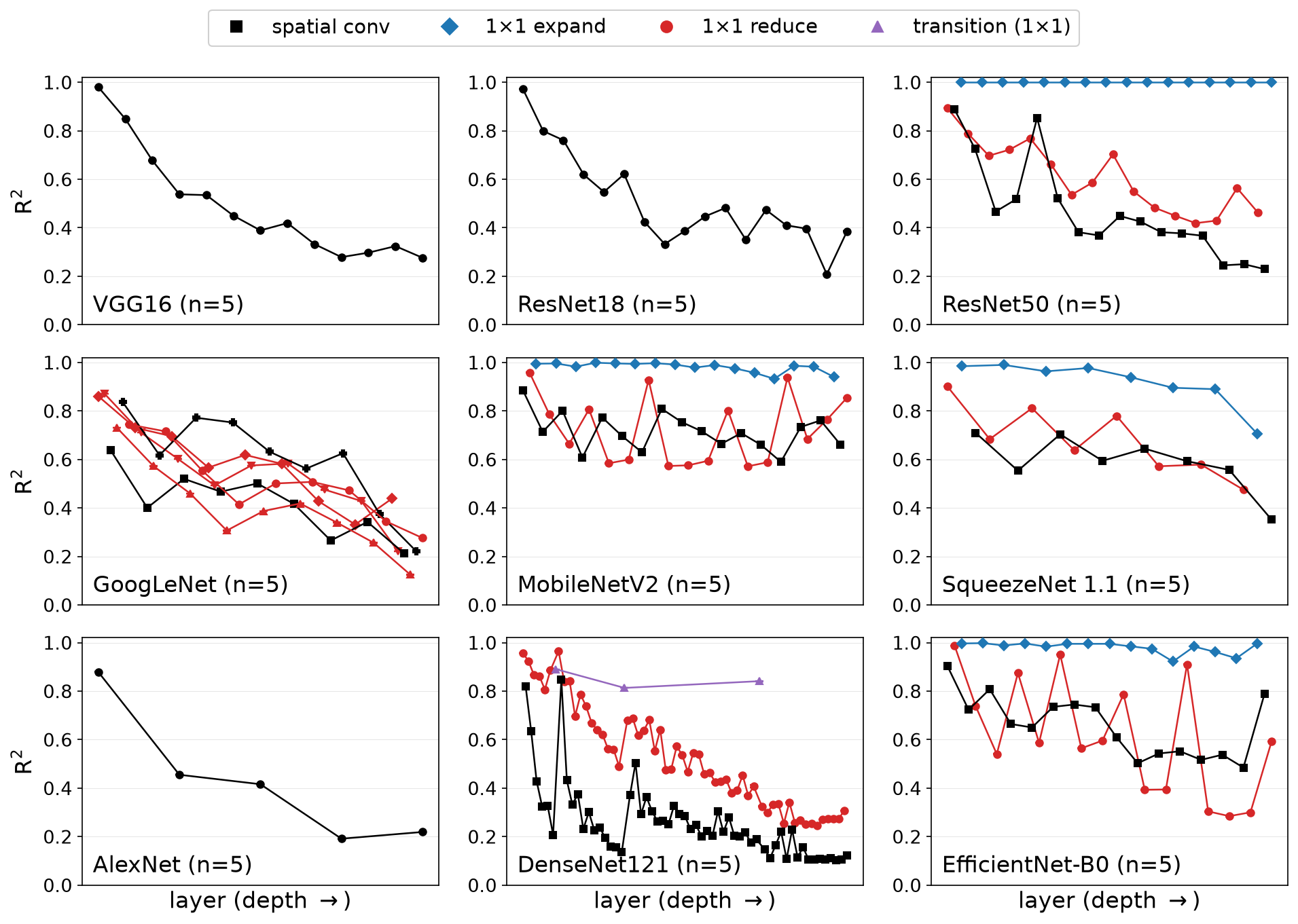}
   \caption{Pretrained recoverability ($R^2_{pre}$) over depth, colored by layer type: convolution where kernel size $K>1$ (black), $1\times1$ channel reduce (red), $1\times1$ channel expand (blue), and DenseNet's transition layers (purple). }
  \label{fig:layer_type_comparison}
 \end{figure}

\begin{figure}[t]
  \centering
  \includegraphics[width=0.92\linewidth]{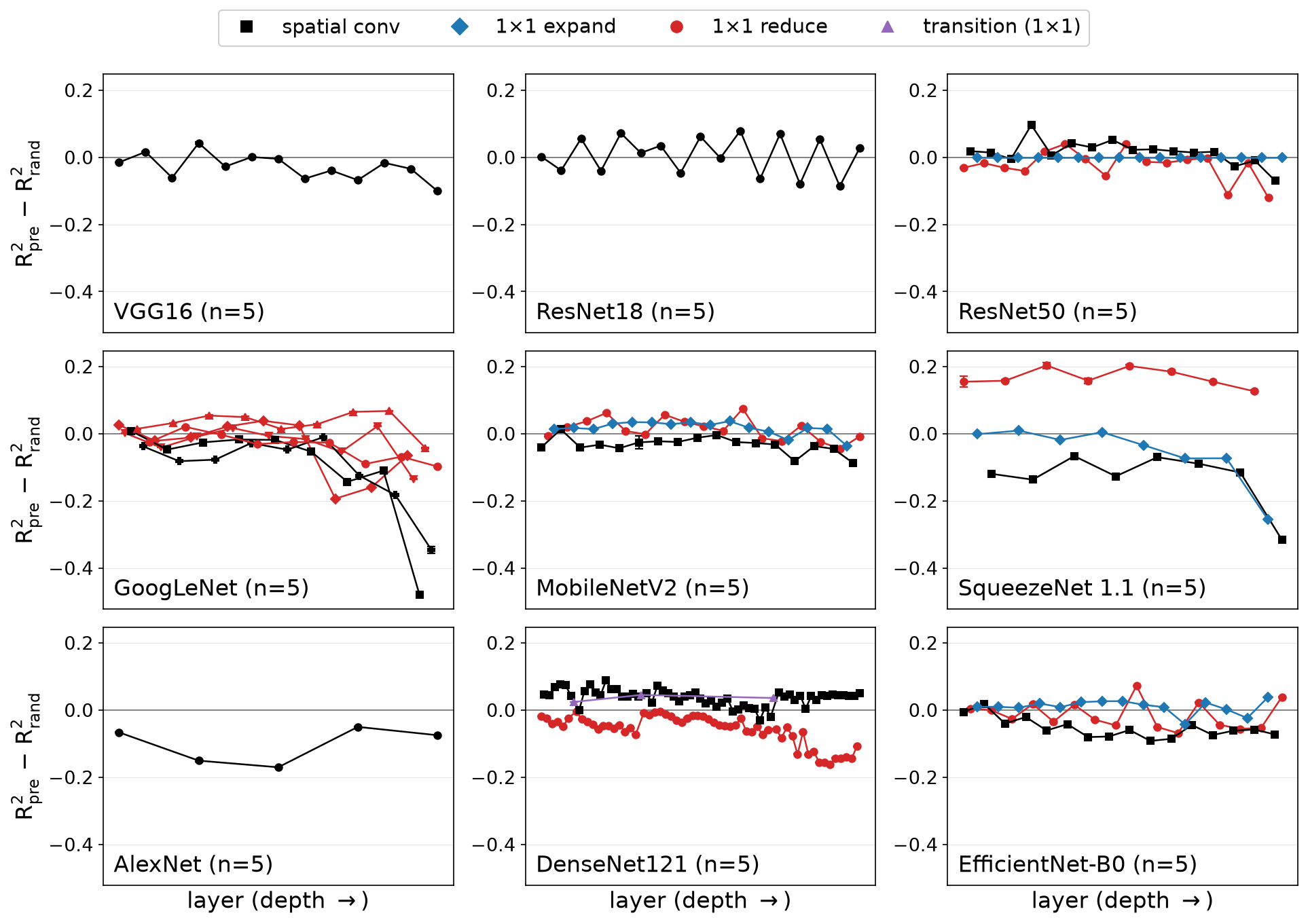}
   \caption{Learning-attributed recoverability, $R^2_{pre}-R^2_{rand}$, over depth for all layers of each architecture, colored by layer type as in Figure~\ref{fig:layer_type_comparison}. Positive values mean the pretrained encoder recovers more of the input than a random encoder of the same shape; the horizontal line marks zero. Error bars are standard errors over probe seeds and all panels share one y-axis.}
  \label{fig:architecture_comparison_gap}
 \end{figure}

\begin{figure}[t]
  \centering
  \includegraphics[width=1\linewidth]{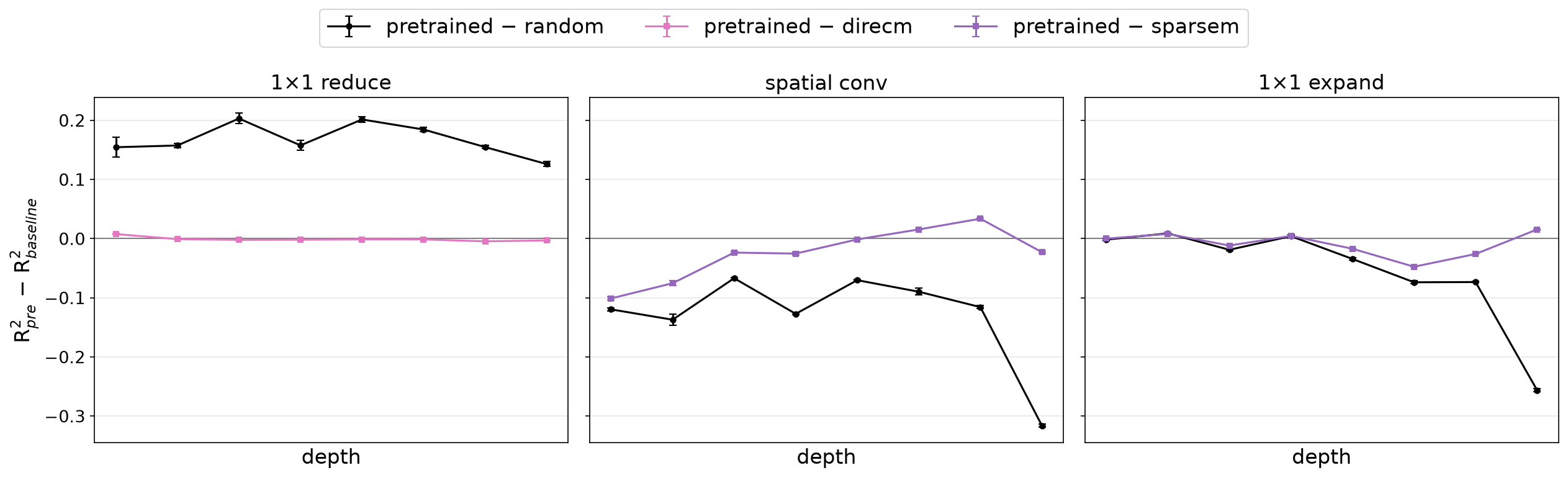}
   \caption{Learning-attributed recoverability and the control that accounts for it in SqueezeNet 1.1, over depth within each layer type. Black: $R^2_{pre}-R^2_{rand}$. Color: the gap that remains after transferring one property, $R^2_{pre}-R^2_{control}$, for the best-scoring control in each layer type (\texttt{direcm} for $1\times1$ reduce layers, \texttt{sparsem} for $K>1$ convolutions and $1\times1$ expand layers); a control that reproduces the pretrained encoder lies on zero. Error bars are standard errors over probe seeds (5 for the random baseline, 3 for the controls). We see that copying pretrained singular vectors accounts for all of the learning-attributed gap in SqueezeNet $1\times1$ reduce layers.}
  \label{fig:gap_by_type_cases_squeezenet}
 \end{figure}

Figure~\ref{fig:layer_type_comparison} shows pretrained recoverability for every probed layer, colored by layer type. Figure~\ref{fig:architecture_comparison_gap} shows learning-attributed recoverability, $R^2_{pre}-R^2_{rand}$, for the same layers, and Figure~\ref{fig:gap_by_type_cases_squeezenet} follows it within each layer type of SqueezeNet 1.1, together with the control that accounts for it best.

\end{document}

%% file: math_commands.tex
\usepackage{amsmath,amsfonts,bm}

\def\eqref#1{equation~\ref{#1}}

\def\1{\bm{1}}

\DeclareMathAlphabet{\mathsfit}{\encodingdefault}{\sfdefault}{m}{sl}
\SetMathAlphabet{\mathsfit}{bold}{\encodingdefault}{\sfdefault}{bx}{n}



%% file: iclr2027_conference.bib
@article{zhaoping2025,
    title   = {Vision: looking and seeing through our brain's information bottleneck},
    author  = {Zhaoping, Li},
    journal = {arXiv preprint arXiv:2503.18804},
    year    = {2025}
}

@ARTICLE{1056814,
  author={Sziklai, G.},
  journal={IRE Transactions on Information Theory}, 
  title={Some studies in the speed of visual perception}, 
  year={1956},
  volume={2},
  number={3},
  pages={125-128},
  doi={10.1109/TIT.1956.1056814}}

@misc{tishby2015deeplearninginformationbottleneck,
      title={Deep Learning and the Information Bottleneck Principle}, 
      author={Naftali Tishby and Noga Zaslavsky},
      year={2015},
      eprint={1503.02406},
      archivePrefix={arXiv},
      primaryClass={cs.LG},
      url={https://arxiv.org/abs/1503.02406}, 
}

@misc{shwartzziv2017openingblackboxdeep,
      title={Opening the Black Box of Deep Neural Networks via Information}, 
      author={Ravid Shwartz-Ziv and Naftali Tishby},
      year={2017},
      eprint={1703.00810},
      archivePrefix={arXiv},
      primaryClass={cs.LG},
      url={https://arxiv.org/abs/1703.00810}, 
}

@inproceedings{
michael2018on,
title={On the Information Bottleneck Theory of Deep Learning},
author={Andrew Michael Saxe and Yamini Bansal and Joel Dapello and Madhu Advani and Artemy Kolchinsky and Brendan Daniel Tracey and David Daniel Cox},
booktitle={International Conference on Learning Representations},
year={2018},
url={https://openreview.net/forum?id=ry_WPG-A-},
}

@misc{yu2020understandingconvolutionalneuralnetworks,
      title={Understanding Convolutional Neural Networks with Information Theory: An Initial Exploration}, 
      author={Shujian Yu and Kristoffer Wickstrøm and Robert Jenssen and Jose C. Principe},
      year={2020},
      eprint={1804.06537},
      archivePrefix={arXiv},
      primaryClass={cs.LG},
      url={https://arxiv.org/abs/1804.06537}, 
}

@article{Gabri__2019,
   title={Entropy and mutual information in models of deep neural networks*},
   volume={2019},
   ISSN={1742-5468},
   url={http://dx.doi.org/10.1088/1742-5468/ab3430},
   DOI={10.1088/1742-5468/ab3430},
   number={12},
   journal={Journal of Statistical Mechanics: Theory and Experiment},
   publisher={IOP Publishing},
   author={Gabrié, Marylou and Manoel, Andre and Luneau, Clément and Barbier, Jean and Macris, Nicolas and Krzakala, Florent and Zdeborová, Lenka},
   year={2019},
   month=Dec, pages={124014} }

@article{9468892,
  author = {Geiger, Bernhard C.},
  journal = {IEEE Transactions on Neural Networks and Learning Systems},
  title = {On Information Plane Analyses of Neural Network Classifiers—A Review},
  year = {2022},
  volume = {33},
  number = {12},
  pages = {7039-7051},
  doi = {10.1109/TNNLS.2021.3089037},
}

@article{e25060899,
  author = {Wickstrøm, Kristoffer K. and Løkse, Sigurd and Kampffmeyer, Michael C. and Yu, Shujian and Príncipe, José C. and Jenssen, Robert},
  title = {Analysis of Deep Convolutional Neural Networks Using Tensor Kernels and Matrix-Based Entropy},
  journal = {Entropy},
  volume = {25},
  year = {2023},
  number = {6},
  article-number = {899},
  url = {https://www.mdpi.com/1099-4300/25/6/899},
  pubmedid = {37372243},
  issn = {1099-4300},
  doi = {10.3390/e25060899},
}

@inbook{Cheng2018,
  title = {Evaluating Capability of Deep Neural Networks for Image Classification via Information Plane},
  isbn = {9783030012526},
  issn = {1611-3349},
  url = {http://dx.doi.org/10.1007/978-3-030-01252-6_11},
  doi = {10.1007/978-3-030-01252-6_11},
  booktitle = {Computer Vision – ECCV 2018},
  publisher = {Springer International Publishing},
  author = {Cheng, Hao and Lian, Dongze and Gao, Shenghua and Geng, Yanlin},
  year = {2018},
  pages = {181–195},
}

@article{Wickstrm2023,
  title = {Analysis of Deep Convolutional Neural Networks Using Tensor Kernels and Matrix-Based Entropy},
  volume = {25},
  issn = {1099-4300},
  url = {http://dx.doi.org/10.3390/e25060899},
  doi = {10.3390/e25060899},
  number = {6},
  journal = {Entropy},
  publisher = {MDPI AG},
  author = {Wickstrøm, Kristoffer K. and Løkse, Sigurd and Kampffmeyer, Michael C. and Yu, Shujian and Príncipe, José C. and Jenssen, Robert},
  year = {2023},
  month = June,
  pages = {899},
}

@article{8594568,
  author = {Shomron, Gil and Weiser, Uri},
  journal = {IEEE Computer Architecture Letters},
  title = {Spatial Correlation and Value Prediction in Convolutional Neural Networks},
  year = {2019},
  volume = {18},
  number = {1},
  pages = {10-13},
  doi = {10.1109/LCA.2018.2890236},
}

@misc{xu2020theoryusableinformationcomputational,
  title = {A Theory of Usable Information Under Computational Constraints},
  author = {Yilun Xu and Shengjia Zhao and Jiaming Song and Russell Stewart and Stefano Ermon},
  year = {2020},
  eprint = {2002.10689},
  archiveprefix = {arXiv},
  primaryclass = {cs.LG},
  url = {https://arxiv.org/abs/2002.10689},
}

@misc{dosovitskiy2016invertingvisualrepresentationsconvolutional,
  title = {Inverting Visual Representations with Convolutional Networks},
  author = {Alexey Dosovitskiy and Thomas Brox},
  year = {2016},
  eprint = {1506.02753},
  archiveprefix = {arXiv},
  primaryclass = {cs.NE},
  url = {https://arxiv.org/abs/1506.02753},
}

@misc{alain2018understandingintermediatelayersusing,
      title={Understanding intermediate layers using linear classifier probes}, 
      author={Guillaume Alain and Yoshua Bengio},
      year={2018},
      eprint={1610.01644},
      archivePrefix={arXiv},
      primaryClass={stat.ML},
      url={https://arxiv.org/abs/1610.01644}, 
}

@misc{zeiler2013visualizingunderstandingconvolutionalnetworks,
      title={Visualizing and Understanding Convolutional Networks}, 
      author={Matthew D Zeiler and Rob Fergus},
      year={2013},
      eprint={1311.2901},
      archivePrefix={arXiv},
      primaryClass={cs.CV},
      url={https://arxiv.org/abs/1311.2901}, 
}

@misc{mahendran2014understandingdeepimagerepresentations,
      title={Understanding Deep Image Representations by Inverting Them}, 
      author={Aravindh Mahendran and Andrea Vedaldi},
      year={2014},
      eprint={1412.0035},
      archivePrefix={arXiv},
      primaryClass={cs.CV},
      url={https://arxiv.org/abs/1412.0035}, 
}

@misc{belinkov2021probingclassifierspromisesshortcomings,
      title={Probing Classifiers: Promises, Shortcomings, and Advances}, 
      author={Yonatan Belinkov},
      year={2021},
      eprint={2102.12452},
      archivePrefix={arXiv},
      primaryClass={cs.CL},
      url={https://arxiv.org/abs/2102.12452}, 
}

@misc{pimentel2020informationtheoreticprobinglinguisticstructure,
      title={Information-Theoretic Probing for Linguistic Structure}, 
      author={Tiago Pimentel and Josef Valvoda and Rowan Hall Maudslay and Ran Zmigrod and Adina Williams and Ryan Cotterell},
      year={2020},
      eprint={2004.03061},
      archivePrefix={arXiv},
      primaryClass={cs.CL},
      url={https://arxiv.org/abs/2004.03061}, 
}
